\documentclass[11pt]{article}
\usepackage[noaddress]{chicagor}

\usepackage{amsmath}
\usepackage{amssymb}
\usepackage{amsfonts}
\usepackage{amsthm}
\usepackage{times}

\allowdisplaybreaks[2]

\newtheoremstyle{theorem}{\topsep}{\topsep}%
     {\itshape}%
     {}%
     {\bfseries}%
     {.}%
     {10pt}%
     {\thmname{#1}\thmnumber{ #2}\thmnote{ (#3)}}%

\theoremstyle{theorem}
\newtheorem{theorem}{Theorem}[section]
\newtheorem{lemma}[theorem]{Lemma}

\newtheorem{EXAMPLE}[theorem]{Example}
\newtheorem{REMARK}[theorem]{Remark}

\newenvironment{example}{\begin{EXAMPLE} \rm}%
                            { \wbox\end{EXAMPLE}}
\newenvironment{example*}{\begin{EXAMPLE} \rm}%
                            {\end{EXAMPLE}}
                            { \wbox\end{REMARK}}
\newenvironment{remark*}{\begin{REMARK} \rm}%
                            {\end{REMARK}}

\newenvironment{oldtheorem}[1]
  {\begin{renewcommand}{\thetheorem}{\ref{#1}}}
  {\end{renewcommand}\addtocounter{theorem}{-1}}

\def\squareforqed{\hbox{\rlap{$\sqcap$}$\sqcup$}}
\def\wbox{\ifmmode\squareforqed\else{\unskip\nobreak\hfil
\penalty50\hskip1em\null\nobreak\hfil\squareforqed
\parfillskip=0pt\finalhyphendemerits=0\endgraf}\fi}

\newcommand{\sat}{\models}
\newcommand{\satS}{\models_{\scriptscriptstyle\rm K}}
\newcommand{\rimp}{\Rightarrow}
\newcommand{\riff}{\Leftrightarrow}
\renewcommand{\phi}{\varphi}

\renewcommand{\emptyset}{\varnothing}

\newcommand{\true}{\textbf{true}}
\newcommand{\false}{\textbf{false}}
\newcommand{\truep}{\mathit{true}}
\newcommand{\falsep}{\mathit{false}}
\newcommand{\Vars}{\mathit{Vars}}
\newcommand{\AX}{{\rm AX}}
\newcommand{\axiom}[1]{\text{#1}}
\newcommand{\mi}[1]{\mathit{#1}}

\newenvironment{prog}{\begin{array}[t]{@{}l@{}}}{\end{array}}

\newcommand{\op}[1]{\mathsf{#1}}
\newcommand{\Ob}{\mathit{Ob}}

\newcommand{\OpAnd}{\mathsf{and}}
\newcommand{\OpKnow}{\mathsf{know}}
\newcommand{\OpDknow}{\mathsf{xknow}}
\newcommand{\OpNot}{\mathsf{not}}
\newcommand{\OpTrue}{\mathsf{true}}
\newcommand{\OpFalse}{\mathsf{false}}
\newcommand{\OpOb}{\mathsf{ob}}

\newcommand{\abs}[1]{\lvert#1\rvert}

\newenvironment{wideitemize}[1]
   {\begin{list}{$\bullet$}
                     {\setlength{\labelwidth}{#1}
                      \setlength{\leftmargin}{#1}
                      \setlength{\itemsep}{0pt}}}
   {\end{list}}
\newenvironment{axiomlist}
   {\begin{wideitemize}{8ex}}
   {\end{wideitemize}}

\newcommand{\LKDprop}{\mathcal{L}^{\scriptscriptstyle\rm KD}}
\newcommand{\SKD}{\Sigma^{\scriptscriptstyle\rm KD}}
\newcommand{\SKDn}{\Sigma^{\scriptscriptstyle\rm KD}_n}

\newcommand{\LKDnprop}{\mathcal{L}^{\scriptscriptstyle\rm KD}_n}
\newcommand{\LK}{\mathcal{L}^{\scriptscriptstyle\rm K}}
\newcommand{\LKn}{\mathcal{L}^{\scriptscriptstyle\rm K}_n}

\newcommand{\C}{\mathcal{C}}
\newcommand{\cM}{\mathcal{M}}
\newcommand{\cK}{\mathcal{K}}
\newcommand{\Axprop}{\AX}
\newcommand{\vdashDY}{\vdash_{\scriptscriptstyle\rm DY}}

\newcommand{\wpf}{\wp_{\scriptscriptstyle\mathit{fin}}}

\newcommand{\COMMENTOUT}[1]{}

\title{Deductive Algorithmic Knowledge\thanks{A preliminary version of 
this paper appeared in the \emph{Proceedings of the 
Eighth International Symposium on Artificial Intelligence and
Mathematics}, AI\&M 22-2004, 2004.  This work was mostly done while the author was at Cornell University.}} 

  \author{Riccardo Pucella\\
  Northeastern University\\
  Boston, MA 02115 USA\\
  riccardo@ccs.neu.edu}

\date{}

\begin{document}
  \maketitle

\begin{abstract}
The framework of algorithmic knowledge assumes that agents use
algorithms to compute the facts they explicitly know. In many cases of
interest, a deductive system, rather than a particular algorithm,
captures the formal reasoning used by the agents to compute what they
explicitly know. We introduce a logic for reasoning about both
implicit and explicit knowledge with the latter defined with respect
to a deductive system formalizing a logical theory for agents. The
highly structured nature of deductive systems leads to very
natural axiomatizations of the resulting logic when interpreted over any
fixed deductive system.  The decision problem for the logic, in the
presence of a single agent,  is
NP-complete in general, no harder than propositional logic.
It remains NP-complete when we fix a deductive system that is
decidable in nondeterministic polynomial time.    
These results extend in a straightforward way to multiple 
agents.
\end{abstract}

\section{Introduction}

It is well known that the standard model of knowledge based on
possible worlds is subject to the problem of \emph{logical
omniscience}, that is, the agents know all the logical consequences of
their knowledge \cite[Chapter 9]{r:fagin95}. 
Thus, possible-world
definitions of knowledge make it difficult to reason about the
knowledge that agents need to explicitly compute in order to make
decisions and perform actions, or to capture situations where agents
want to reason about the knowledge that other agents need to
explicitly compute in order to perform actions.

This observation leads to a distinction between two forms of knowledge, 
\emph{implicit knowledge} and \emph{explicit
knowledge} (or resource-bounded knowledge), a distinction long 
recognized \cite{r:rosenschein85}. The classical AI approach known as
the \emph{interpreted symbolic structures} approach, where knowledge
is based on information stored in data structures of the agent, can be
seen as an instance of explicit knowledge. In contrast, the
\emph{situated automata} approach, which interprets knowledge based on 
information carried by the state of the machine, can be seen as an
instance of implicit knowledge. Levesque \citeyear{r:levesque84} makes
a similar distinction between implicit belief and explicit belief.

While the possible-worlds approach is taken as the standard model for
implicit knowledge, there is no standard model for
explicit knowledge.  A general approach appropriate for many
situations is that of \emph{algorithmic knowledge} \cite{r:halpern94}.
In the algorithmic knowledge framework, the explicit knowledge of an
agent is given by a knowledge algorithm that the agent uses to
establish whether he knows a particular fact.  Algorithmic knowledge
is sufficiently expressive to capture a number of approaches to
resource-bounded reasoning that have appeared in the literature
\cite{r:levesque84,r:konolige86,ElgPerlis}.%
\footnote{While we focus on knowledge in this paper, much of our
development applies equally well to belief. In fact, since we do not
assume that knowledge algorithms necessarily return the correct
answer, one could argue that the kind of knowledge provided by
knowledge algorithms is really belief. For consistency with the
literature on algorithmic knowledge, however, we shall continue to use
the terminology ``algorithmic knowledge''.}

The generality of the algorithmic knowledge approach makes it ideal as
a modeling framework. One consequence of that generality, however, is
that there are no nontrivial logical properties of algorithmic knowledge
proper, unless we focus on specific classes of knowledge
algorithms. This consequence is important if the framework is to be
used as a specification language for properties of multiagent systems,
amenable to automated verification. In such a setting, we would like a
class of knowledge algorithms that can capture properties of interest
for the verification task at hand, while retaining enough structure to
yield a tractable or analyzable framework. This structure will typically
implies a number of properties of the corresponding algorithmic
knowledge operator, which can be used to study properties of
multiagent systems purely deductively. This general observation leads
naturally to a program of studying interesting classes of knowledge
algorithms.

In this paper, we study a form of algorithmic knowledge,
\emph{deductive algorithmic knowledge}, where the explicit knowledge
of agents comes from a logical theory expressed by a deductive system
made up of deduction rules, in which the agents perform their
reasoning about the facts they know.  
Many useful forms of explicit knowledge can be formalized using
deductive systems.
For instance, Horn theories \cite{r:selman96}, which have been used to
approximate more general knowledge bases, fit into this framework.
Explicit knowledge via a deductive system can be viewed as a form of
algorithmic knowledge, where the knowledge algorithm used by
an agent attempts to infer whether a fact is
derivable from the deduction rules provided by the agent's deductive
system.
Among other advantages, viewing deductive algorithmic knowledge as an
instance of algorithmic knowledge lets us model ``feasible'' explicit
knowledge, by considering deductive systems whose corresponding
knowledge algorithm is efficient (e.g., runs in polynomial time).
The approach to modeling explicit knowledge through deductive systems
is essentially that of Konolige \citeyear{r:konolige86}. The basic
idea, which we review in Section~\ref{s:deductivesystems}, is that a
deductive system is a set of rules that describe how to infer new
facts from old facts. We take the initial facts of an agent to be his
observations, that is, what the agent can determine simply by
examining his local state.

We describe in Section~\ref{s:deductive} a logic for reasoning about
both implicit knowledge and deductive algorithmic knowledge, with
formulas $K\phi$ to express implicit knowledge of $\phi$, and $X\phi$
to express deductive algorithmic knowledge of $\phi$. For simplicity,
the logic we present is propositional, although there is no difficulty
in extending it to a first-order setting.
(Of course, the standard questions about interactions between
quantification and modal operators arise in such a setting.)
Why reason about two forms of
knowledge? It turns out that in many situations, one wants to reason
about both forms of knowledge. Intuitively, implicit knowledge is
useful for specifications and describes ``ideal'' knowledge, while
deductive algorithmic knowledge is useful to capture the knowledge that
agents can actually compute and use. Consider the following
example. In previous work, we showed how the framework of algorithmic
knowledge could be used to reason about agents communicating through
cryptographic protocols
\cite{r:halpern02e}. Algorithmic knowledge is useful to model an
adversary that has certain capabilities for decoding the messages he
intercepts. There are of course restrictions on the capabilities of a
reasonable adversary. For instance, the adversary may not explicitly
know that he has a given message if that message is encrypted using a
key that the adversary does not know. To capture these restrictions,
Dolev and Yao \citeyear{r:dolev83} gave a now-standard description of
capabilities of adversaries.  Roughly speaking, a Dolev-Yao adversary
can decompose messages, or decipher them if he knows the right keys,
but cannot otherwise ``crack'' encrypted messages. The adversary can
also construct new messages by concatenating known messages, or
encrypting them with a known encryption key.  It is natural to
formalize a Dolev-Yao adversary using a deductive system that
describes what messages the adversary possesses based on the messages he
has intercepted, and what messages the adversary can construct. This
lets us describe properties such as ``the adversary can compute (i.e.,
explicitly knows) the secret exchanged during the protocol run''. To
capture the fact that some other agent in the system knows that a
secret exchanged during the protocol run in fact remains a secret from
the adversary, we can use implicit knowledge, as in ``agent $A$ knows
(i.e., implicitly knows) that the adversary cannot compute (i.e.,
explicitly knows) the secret exchanged during the protocol run''. In
this sense, specifications can refer to both kinds of knowledge. (The
specification above requires an extension of the logic to
handle multiple agents; see Section~\ref{s:multiagent}.)

A key operator in the logic is the operator $\Ob$ that 
identifies the observations made at a state. In a precise sense, this
operator is the connection between implicit knowledge and explicit
knowledge: deductive algorithmic knowledge uses observations as the
initial facts from which further facts explicitly known are derived,
while implicit knowledge uses observations to distinguish states, in
that two states are indistinguishable exactly when the agent has the
same observations in both states. Thus, we can study the interaction
between implicit and explicit knowledge, and go beyond previous work
that only attempts to model explicit knowledge
\cite{r:konolige86,r:giunchiglia93}.  Our work  shows one way
to combine a standard possible-worlds account of implicit knowledge
with a deductive system representing the explicit knowledge of agents,
and to reason about both simultaneously.

A principal goal of this paper is to study the technical properties of
the resulting logic, such as axiomatizations and complexity of 
decision problems, and see how they relate to properties of the
deductive systems. In Section~\ref{s:axiomatizations}, we study
axiomatizations for reasoning about
specific deductive systems. Not surprisingly, if we do not make any
assumption on the deductive system, there are very few properties
captured by the axiomatization, which is essentially a slight
extension of well-known axiomatizations for implicit knowledge. However, if we
restrict our attention to models where the agent uses a specific
deductive system, then the properties of $X$  depend on that
deductive system. Intuitively, we should be able to read off the
properties of $X$ from the deduction rules.  We formalize
this intuition by showing that we can derive sound and complete
axiomatizations for our logic with respect to models equipped with a
specific deductive system, where the axiomatization is derived
mechanically from the rules of the deductive system.  

In Section~\ref{s:complexity}, we address the complexity of the
decision problem for the logic in the presence of a single agent, that
is, the complexity of deciding if a formula of the logic is
satisfiable with respect to a class of 
models. Without any assumption on the deductive systems, deciding
satisfiability is NP-complete; this is not surprising, since the logic
in that context is essentially just a logic of implicit knowledge,
known to be NP-complete \cite{r:ladner77} when 
there is a single agent. 
If we fix a
specific deductive system, then deciding satisfiability with respect
to the class of models using that deductive system is NP-hard, and it
is possible to show NP-completeness when the deductive
system is itself decidable in nondeterministic polynomial time and the
models considered 
have a small number of observations at each state.

In Section~\ref{s:multiagent}, we consider a natural extension of our
framework to reason about multiple agents.  This is needed to study
the Dolev-Yao example given above in its full generality, or any
interesting example from the multiagent systems literature.  The
extension is straightforward, and many of the results generalize in
the obvious way, justifying our decision to focus on the single agent
model for the bulk of the paper.
For instance, the complexity of the decision problem when there are
multiple agents becomes PSPACE-complete, again following from 
logics of implicit knowledge themselves having PSPACE-complete
decision problems \cite{r:halpern92} in the presence of multiple
agents.  
The proofs of our technical results are deferred to the appendices.

\section{Deductive Systems}\label{s:deductivesystems}

We start by defining the framework in which we capture the logical
theories of the agents, that is, their deductive or inferential
powers.  
We distinguish the logical theories of the agents from the logic that
we introduce in the next section to reason about a system and
what agents in the system know. 
We can therefore model agents with different inferential powers,
without affecting the logic itself.

Following common practice, we take deductive systems as
acting over the terms of some term algebra.
More precisely, assume a fixed finite signature
$\Sigma=(f_1,\ldots,f_n)$, where each $f_i$ is an operation symbol,
with arity $r_i$. Operation symbols of arity $0$ are called constants.
Assume a countable set $\Vars$ of variables. Define the \emph{term
algebra} $T_\Sigma$ as the least set such that $\Vars\subseteq
T_\Sigma$, and for all $f\in\Sigma$ of arity $n$, and for all
$t_1,\ldots,t_n\in T_\Sigma$, then $f(t_1,\ldots,t_n)\in
T_\Sigma$. Intuitively, $T_\Sigma$ contains all the terms that can be
built from the variables, constants, and operations in $\Sigma$. We
say a term is a
\emph{ground term} if it contains no variables. Let $T^g_\Sigma$ be
the set of ground terms in $T_\Sigma$. A 
ground
substitution $\rho$ is a mapping from variables in $\Vars$ to ground
terms. The application of a ground substitution $\rho$ to a term $t$, written
$\rho(t)$, essentially consists of replacing every variable in $t$
with the ground term corresponding to $t$ in $\rho$. Clearly, the
application of a ground substitution to a term yields a ground term.

A \emph{$\Sigma$-deductive system} $D$ is a subset of $\wpf(T_\Sigma)\times
T_\Sigma$. (We write $\wp(X)$ for the set of subsets of $X$, and
$\wpf(X)$ for the set of finite subsets of $X$.) We often omit the
signature $\Sigma$ when it is clear from context. A \emph{deduction
rule} $(\{t_1,\ldots,t_n\},t)$ of $D$ is typically written
$t_1,\ldots,t_n\triangleright t$, and means that $t$ can be
immediately deduced from $t_1,\ldots,t_n$. A deduction of $t$ from a
set $\Gamma$ of terms is a sequence of ground terms $t_1,\ldots,t_n$
such that $t_n=t$, and every $t_i$ is either
\begin{enumerate}
\item a term $\rho(t')$, for some ground substitution $\rho$ and some term $t'\in\Gamma$;
\item a term $\rho(t')$, for some ground substitution $\rho$ and some
term $t'$ for which there is a deduction rule
$t'_{i_1},\ldots,t'_{i_k}\triangleright t'$ in $D$ such that
$\rho(t'_{i_j})=t_{i_j}$ for all $j$, and $i_1,\ldots,i_j<i$.
\end{enumerate}
We
write $\Gamma\vdash_D t$ if there is a deduction from $\Gamma$ to $t$
via deduction rules in $D$. By definition, we have $t
\vdash_D t$ for all terms $t$.%
\footnote{Our use of variables in deductive systems is purely for
convenience. We could replace every deduction rule with the rules
obtained by substituting ground terms for all the variables, and
get a deductive system that can derive the same terms as the
original one.}

We will mainly be concerned with deductive systems that are
\emph{decidable}, that is, for which the problem of deciding whether a 
deduction of $t$ from $\Gamma$ exists is decidable, for a term $t$ and
a finite set of terms $\Gamma$. Moreover, it should be clear from the
definitions that deductive systems are
monotonic. In other words, if $\Gamma\vdash_D t$, then
$\Gamma'\vdash_D t$ when $\Gamma\subseteq\Gamma'$. Finally, observe
that we do not impose any restriction on the formation of terms. The
whole theory we develop in this paper could take as a starting point
the notion of \emph{sorted term algebras} \cite{r:higgins63}, with
little change; this would allow restrictions on terms to be imposed in
a natural way.

\begin{example}\label{x:dy} 
We give a deductive system that captures the capabilities of the
Dolev-Yao adversary described in the introduction. Define the following
$\Sigma$-deductive system DY, with signature
$\Sigma=(\op{recv},\op{has},\op{encr},\op{conc},\op{inv})$, where
$\op{recv}(m)$ represents the fact that the adversary has received the
term $m$, $\op{has}(m)$ represents the fact that the adversary
possesses the term $m$ (i.e., is able to extract message $m$ from the
messages he has received), $\op{encr}(m,k)$ represents the encryption
of term $m$ with key $k$, $\op{conc}(m_1,m_2)$ represents the
concatenation of terms $m_1$ and $m_2$, and $\op{inv}(k)$ represents
the inverse of the key $k$:
\[\begin{array}{rcl}
\op{recv}(m) & \triangleright & \op{has}(m)\\
\op{has}(\op{inv}(k)),\op{has}(\op{encr}(m,k)) & \triangleright & \op{has}(m)\\
\op{has}(\op{conc}(m_1,m_2)) & \triangleright & \op{has}(m_1)\\
\op{has}(\op{conc}(m_1,m_2)) & \triangleright & \op{has}(m_2).
  \end{array}\]
Assume further that $\Sigma$ contains constants such as
$\op{m},\op{k}_1,\op{k}_2$. We can  derive:
\[\op{recv}(\op{encr}(\op{m},\op{k}_1)), 
\op{recv}(\op{encr}(\op{inv}(\op{k}_1),\op{k}_2)),
\op{recv}(\op{inv}(\op{k}_2)) \vdashDY \op{has}(\op{m}).\]
In other words, it is possible for a Dolev-Yao adversary to derive the
message $\op{m}$ if he has received $\op{m}$ encrypted under a key
$\op{k}_1$, the inverse of which he has received encrypted under a key
$\op{k}_2$, whose inverse he has received.

To account for constructing new messages, consider the signature
$\Sigma'$ that extends $\Sigma$ with a unary constructor
$\op{constr}$, where $\op{constr}(m)$ represents the fact that the
adversary can construct the term $m$. We can account for this new
constructor by adding the following deduction rules to DY:
\[\begin{array}{rcl}
\op{has}(m) & \triangleright & \op{constr}(m)\\
\op{constr}(k),\op{constr}(m) & \triangleright & \op{constr}(\op{encr}(m,k))\\
\op{constr}(m_1), \op{constr}(m_2) & \triangleright &
\op{constr}(\op{conc}(m_1,m_2)).
  \end{array}\]
For instance, we have
\[\op{recv}(\op{encr}(\op{m},\op{k}_1)),
\op{recv}(\op{inv}(\op{k}_1)), \op{recv}(\op{k}_2) 
\vdashDY \op{constr}(\op{encr}(\op{m},\op{k}_2)).\]
\end{example}

In what sense can we use deductive systems to model explicit
knowledge? Intuitively, the elements of the term algebra represent
facts, and the deduction rules of the system model the inferencing
capabilities of the agent---which facts can he deduce from other
facts. This gloss raises another question, namely what to take as basic
facts known to the agent without deduction, from which to initially
start deriving other facts? Konolige \citeyear{r:konolige86} calls these
\emph{basic beliefs}, and there are a number of approaches that can be
followed. One approach is to simply posit a set of basic facts initially
known to the agent (perhaps different basic facts are initially known
at different states). This does not completely solve the problem, as
it still requires determining which facts are initially known at every
state. We pursue  a different approach here, consistent with many
well-known descriptions in the literature: we take the basic beliefs
of an agent to be his
observations. Intuitively, an observation is a fact that the agent can
readily determine by examining his local state. We distinguish
observations from other facts by using a unary constructor
$\OpOb$ in the signature of the deductive systems we consider in subsequent
sections.

\section{Deductive Algorithmic Knowledge}\label{s:deductive}

We now introduce a propositional modal logic for reasoning about the
implicit and explicit knowledge of an agent, where the explicit
knowledge is formalized as a deductive system.  In this section, we
focus on a single agent.  We extend to multiple agents in
Section~\ref{s:multiagent}.

We define the logic $\LKDprop(\Sigma)$, over a signature $\Sigma$. We
take primitive propositions to be ground terms $T^g_\Sigma$
over the signature $\Sigma$. We use $p$ to range over $T^g_\Sigma$, to
emphasize that they are primitive propositions, and distinguish them
from terms over the more general signatures described later.  The
language of the logic is obtained by starting with primitive
propositions in $T^g_\Sigma$
and closing off under negation, conjunction, the $K$ operator, the $X$
operator, and the $\Ob$ operator (applied to primitive propositions
only). Intuitively, $K\phi$ is read as ``the agent implicitly knows
$\phi$'', $X \phi$ is read as ``the agent explicitly knows
$\phi$, according to his deductive system'', and $\Ob(p)$ is read as
``the agent observes $p$''.  We define the usual abbreviations,
$\phi\lor\psi$ for $\neg(\neg\phi\land\neg\psi)$, and $\phi\rimp\psi$
for $\neg\phi\lor\psi$.  We define $\truep$ as an abbreviation for an
arbitrary but fixed propositional tautology, and $\falsep$ as an
abbreviation for $\neg\truep$.

To interpret the deductive algorithmic knowledge of an agent, we
provide the agent with a deductive system in which to perform his
deductions. As we discussed in last section, we want the agent
to reason about observations he makes about his state;
furthermore, because we will want to interpret formulas in $\LKDprop$ as
terms over which the deductive system can reason (see the semantics of
$X\phi$ below), we consider deductive systems defined
over the signature $\Sigma$ extended with a set $\SKD$ of constructors
corresponding to the operators in our logic, that is,
$\SKD=\{\OpOb,\OpTrue,\OpFalse,\OpNot,\OpAnd,\OpKnow,\OpDknow\}$,
where $\OpTrue,\OpFalse$ have arity 0, $\OpOb,\OpNot,\OpKnow,\OpDknow$
have arity 1, and $\OpAnd$ has arity $2$.

\newcommand{\SigmaKD}{\Sigma\mathop{\cup}\SKD}
\newcommand{\SigmaKDn}{\Sigma\mathop{\cup}\SKDn}

The semantics of the logic follows the standard possible-worlds
presentation for modal logics of knowledge \cite{r:hintikka62}.  A
\emph{deductive algorithmic knowledge structure} is a tuple
$M=(S,\pi,D)$, where $S$ is a set of states, $\pi$ is an
interpretation for the primitive propositions, and $D$ is a
$\SigmaKD$-deductive system. Every state $s$ in $S$ is of the form
$(e,O)$, where $e$  captures the general state of the system, and $O$ is a finite
set of observations. Each observation in $O$ is a
primitive proposition, representing the observations that the agent
has made at that state.\footnote{For simplicity, we assume that the observations
form a set. This implies that repetition of observations and their
order is unimportant. We can easily model the case where the
observations form a sequence, at the cost of complicating the
presentation. We also assume, again for simplicity, that there are
only finitely many observations at every state. Allowing infinitely
many observations does not affect anything in this section; we
conjecture that the results in subsequent sections also hold when
infinitely many observations per state are allowed.} 
The details of the states are essentially
irrelevant, as they are only used as a way to interpret the truth of
the primitive propositions. We do not model how
agent makes  observations, or temporal relationships
between states. A state simply represents a snapshot of the system
under consideration.  The interpretation $\pi$ associates with every
state the set of primitive propositions that are true at that state,
so that for every primitive proposition $p\in T^g_\Sigma$, we have
$\pi(s)(p)\in\{\true,\false\}$.

We make a distinction between a fact, represented by a primitive
proposition $p$, and an observation of that fact, represented by the
formula $\Ob(p)$. For instance, the fact that Alice holds an apple
might be represented by the primitive proposition
$\op{holds}(\op{alice},\op{apple})$, which can be true or not at a
state, while the fact that the agent has observed that Alice is
holding an apple is represented by the formula
$\Ob(\op{holds}(\op{alice},\op{apple}))$, which is true if and only if
that observation is in the state of the agent. It is not necessarily
the case that if $\Ob(p)$ holds at a state then $p$ holds at that
state. 
We therefore consider it possible that the agent makes
unreliable observations. We can of course assume that observations are
reliable by imposing a restriction on the interpretation $\pi$, by
taking $\pi((e,O))(p)=\true$ whenever $p\in O$. More generally,
we can impose  restrictions on the models considered, such as
observations being restricted to a specific subset of the primitive
propositions, and so on.%
\footnote{
We can also impose restrictions on deductive systems and
signatures. In a preliminary version of this paper \cite{r:pucella04}, 
for instance, we distinguished the notion of a primitive signature,
which does not provide constructors for the propositional and modal
connectives. A deductive system based on a primitive
signature only permits reasoning about explicit knowledge of
primitive propositions.
}

\begin{example}\label{x:dy3}
Models are representations of situations we want to analyze, for
instance, by  verifying that a  situation satisfies a
certain property. Suppose we wanted to specify the knowledge of a
Dolev-Yao adversary, as mentioned in the introduction, in the context
where there are principals exchanging messages according to a
protocol. The deductive system used by the adversary is a slight
extension of the deductive system DY from Example~\ref{x:dy}, extended
to deal with observations.
The security literature generally assumes a subterm relation on the
messages exchanged by the protocol; define $\sqsubseteq$ on
$T^g_{\scriptscriptstyle\rm DY}$ as the smallest relation satisfying
\[\begin{array}{l}
  t\sqsubseteq t\\ \text{if $t\sqsubseteq t_1$ then $t\sqsubseteq
  \op{conc}(t_1,t_2)$}\\ \text{if $t\sqsubseteq t_2$ then
  $t\sqsubseteq \op{conc}(t_1,t_2)$}\\ \text{if $t\sqsubseteq t_1$
  then $t\sqsubseteq \op{encr}(t_1,t_2)$.}  \end{array}\] 
(See Abadi and Rogaway \citeyear{r:abadi00} for motivations.)
Consider a structure $M=(S,\pi,{\rm DY'})$, where we record at every
state all messages intercepted by the adversary at that state. We
restrict the observations at a state to be of the form $\op{recv}(t)$,
for ground terms $t$ in which $\op{has}$ does not occur. The deductive
system ${\rm DY}'$ is just DY extended with a rule making observations
available to the deductive system:
\[\OpOb(\op{recv}(t))\triangleright \op{recv}(t).\] The
interpretation $\pi$ is defined so that
$\pi((e,O))(\op{has}(t))=\true$ if and only if there exists a term
$t'\in T^g_{\scriptscriptstyle\rm DY}$ such that
$\op{recv}(t')\in O$ and $t\sqsubseteq t'$. In other words,
$\op{has}(t)$ holds at a state if $t$ is a subterm of a message
intercepted by the adversary. For instance, we can have $s_1$ be a
state with observations
\[\{\op{recv}(\op{encr}(\op{m},\op{k}_1)),
\op{recv}(\op{encr}(\op{inv}(\op{k}_1),\op{k}_2))\},\] and $s_2$ a state with
observations 
\[\{\op{recv}(\op{encr}(\op{m},\op{k}_1)),
\op{recv}(\op{encr}(\op{inv}(\op{k}_1),\op{k}_2)),
\op{recv}(\op{inv}(\op{k}_2))\}.\] 
These states represent states where the adversary has intercepted
particular messages. We shall see how to specify properties of $M$,
such as the fact the adversary does not explicitly know at $s_1$ that
he possesses $\op{m}$, despite $\pi(s_1)(\op{has}(\op{m}))=\true$.
\end{example}

Let $\cM(\Sigma)$ be the set of all deductive algorithmic knowledge
structures using $\SigmaKD$-deductive systems. For a fixed
$\SigmaKD$-deductive system $D$, let $\cM_D(\Sigma)$ be the set of
all deductive algorithmic knowledge structures using deductive system
$D$.

We define what it means for a formula $\phi$ to
be true at a state $s$ of $M$, written $(M,s)\sat\phi$, inductively as
follows. For the propositional fragment of the logic, the rules are
straightforward. 
\begin{itemize}
\item[] $(M,s)\sat p$ if $\pi(s)(p)=\true$
\item[] $(M,s)\sat \neg\phi$ if $(M,s)\not\sat\phi$
\item[] $(M,s)\sat \phi\land\psi$ if $(M,s)\sat\phi$ and $(M,s)\sat\psi$.
\end{itemize}
To define the semantics of knowledge, we follow the standard
approach due to Hintikka \citeyear{r:hintikka62}. We define a relation
on the states that captures the states that the agent cannot
distinguish based on the observations. Let $s\sim s'$ if and only if 
$s=(e,O)$ and $s'=(e',O)$ for some $e,e'$, and set of
observations $O$. Clearly, $\sim$ is an equivalence relation on the
states.
\begin{itemize}
\item[] $(M,s)\sat K\phi$ if $(M,s')\sat\phi$ for all $s'\sim s$.
\end{itemize}
To define the semantics of the $X$ operator, we need to
invoke the deductive system.  To do this, we first define the
translation of a formula $\phi$ of $\LKDprop(\Sigma)$ into a term $\phi^T$
of the term algebra, in the completely obvious way: $p^T$ is $p$ for
any primitive proposition $p$ (recall that primitive propositions are
just terms in $T^g_\Sigma$),
$(\neg\phi)^T$ is $\OpNot(\phi^T)$,
$(\phi\land\psi)^T$ is $\OpAnd(\phi^T,\psi^T)$, $(K\phi)^T$ is
$\op{know}(\phi^T)$, $(X\phi)^T$ is $\OpDknow(\phi^T)$, and
$(\Ob(p))^T$ is simply $\OpOb(p)$. 
\begin{itemize}
\item[] $(M,s)\sat X\phi$ if $s=(e,O)$ and $\{\OpOb(p)\mid p\in O\}\vdash_D \phi^T$. 
\end{itemize}
The monotonicity of the deductive systems means that for a structure
$M$ with states $s=(e,O)$, $s'=(e',O')$, and
$O\subseteq O'$, we have $(M,s)\sat X\phi$ implies $(M,s')\sat
X\phi$. Thus, explicit knowledge of facts is never lost when new
observations are made. Finally, we interpret $\Ob(p)$ by checking
whether $p$ is one of the observations made by the agent:
\begin{itemize}
\item[] $(M,s)\sat \Ob(p)$ if $s=(e,O)$ and $p\in O$. 
\end{itemize}
As usual, we say a formula $\phi$ is
\emph{valid in $M$} if $(M,s)\sat\phi$ for all $s\in S$, and
\emph{satisfiable in $M$} if $(M,s)\sat\phi$ for some $s\in S$. If $\cM$ is a
set of models, we say a formula $\phi$ is \emph{valid in $\cM$} if
$\phi$ is valid in every $M\in\cM$, and \emph{satisfiable in
$\cM$} if $\phi$ is satisfiable in some $M\in\cM$. A formula $\phi$ in
$\LKDprop(\Sigma)$ is \emph{valid} if it is valid in
$\cM(\Sigma)$, and \emph{satisfiable} if it is satisfiable in
$\cM(\Sigma)$.

\begin{example}
Consider Example~\ref{x:dy3}.  By definition of $\pi$, $(M,s_1)\sat
K(\op{has}(\op{m}))$ and $(M,s_2)\sat K(\op{has}(\op{m}))$, so that at
both states, the adversary implicitly knows he possesses message
$\op{m}$. However, from the results of Example~\ref{x:dy}, we see that
$(M,s_2)\sat X(\op{has}(\op{m}))$, while $(M,s_1)\sat\neg
X(\op{has}(\op{m}))$. In other words, the adversary explicitly knows
he possesses $\op{m}$ at state $s_2$ (where he has intercepted the
appropriate terms), but not at state $s_1$.
\end{example}

\begin{example}\label{x:bool}
The following deduction rules can be added to any deductive
system to obtain a deductive system that captures a
subset of the inferences that can be performed in propositional logic:
\[\begin{array}{llllll}
  t  \triangleright  \op{not}(\op{not}(t)) & &
    \op{not}(\op{and}(t,\op{not}(t'))), t  \triangleright  t' & & 
      \op{and}(t,t')  \triangleright  t'\\
  \op{not}(\op{not}(t))  \triangleright  t & &
    \op{not}(\op{and}(t,\op{not}(t'))), \op{not}(t')  \triangleright  \op{not}(t) & &
      t, \op{not}(t)  \triangleright  \op{false}\\
  t  \triangleright  \op{not}(\op{and}(\op{not}(t),\op{not}(t'))) & &
    t, t'  \triangleright  \op{and}(t,t') & & 
      \op{false}  \triangleright  t\\
  t'  \triangleright  \op{not}(\op{and}(\op{not}(t),\op{not}(t'))) & &
    \op{and}(t,t')  \triangleright  t. & & \\
  \end{array}\]
One advantage of these rules, despite the fact that they are
incomplete, is that they can be used to perform very efficient
(linear-time, in fact) propositional inference \cite{r:mcallester93}.
\end{example}

\begin{example}\label{x:new1} 
We can easily let the agent explicitly reason about his
deductive algorithmic knowledge by adding a rule 
\begin{equation}
\label{e:selfdknow}
t\triangleright\OpDknow(t)
\end{equation}
to his deductive system $D$. Thus, if $M$ is a deductive algorithmic
knowledge structure over $D$, and $(M,s)\sat X\phi$, then we have
$s=(e,O)$, with $O\vdash_D \phi^T$, and by the above rule, the
deductive system $D$ can also derive $O\vdash_D\OpDknow(\phi^T)$,
so that $O\vdash_D(X\phi)^T$. Thus, $(M,s)\sat X(X\phi)$, as
required. It is possible to restrict the deductive algorithmic
knowledge of an agent with respect to his own deductive algorithmic
knowledge by suitably modifying rule (\ref{e:selfdknow}), restricting
it to apply only to a subset of the terms.
\end{example}

There is a subtlety involved in any logic with a modal operator that
we wish not to be subject to logical omniscience. Intuitively, such a logic
forces one to be quite aware of what symbols are defined by
abbreviation, and which are not.  Earlier in this section, we defined
$\truep$, $\falsep$, $\lor$, and $\rimp$ by abbreviation, which means
that any formula containing $\lor$ or $\rimp$ is really a formula
containing $\land$ and $\neg$. Thus, the agent cannot explicitly
distinguish between $\phi\lor\psi$ and $\neg(\neg\phi\land\neg\psi)$;
they are the same formula in the logic, and the validity $\sat
X(\phi\lor\psi)\riff X(\neg(\neg\phi\land\neg\psi))$ simply reflects
this identity. Such a result seems to go against the main motivation
for explicit knowledge, to ensure that knowledge is not closed under
tautologies. Part of the problem here is simply that defining
operators by abbreviation introduces inescapable equivalences. An easy
way to circumvent this problem is to use a syntax that directly uses
$\lor$, $\rimp$, and perhaps other connectives, rather than
introducing them through abbreviations. We would then add similar
constructors to the signature $\SKD$, and extend the translation
$\phi^T$ accordingly. This gives full control on which tautologies are
validated by explicit knowledge, and which are not. 

Finally, it is worth pointing out the relationship between our
framework and Konolige's \citeyear{r:konolige86}. 
Roughly speaking, Konolige's framework corresponds to the
deductive systems we described in the last section---it supplies a
logical theory for the reasoning of an agent (albeit, in Konolige's
case, a first-order logical theory). 
The logic in this section may be used to reason about the knowledge of
agents who reason using Konolige's logical theories.  
In this sense, our logic is compatible with Konolige's framework.
More interestingly, our framework semantically grounds the basic
beliefs assumed by Konolige's framework: basic beliefs correspond to
observations, which can be reasoned about independently of the belief
of the agents.

\section{Axiomatizations}\label{s:axiomatizations}

In this section, we present a sound and complete axiomatization for
reasoning about explicit knowledge given by a deductive system. Recall
that a formula $f$ is \emph{provable in an axiomatization} if $f$ can
be proved using the axioms and rules of inference of the
axiomatization. An axiomatization is \emph{sound} with respect to a
class $\cM$ of structures if every formula provable in the
axiomatization is valid in $\cM$; an axiomatization is \emph{complete}
with respect to $\cM$ if every formula valid in $\cM$ is provable in
the axiomatization.

Clearly, for a fixed deductive system, the properties of $X$
depend on that deductive system. Intuitively, we should be able to
read off the properties of $X$ from the deduction rules themselves.
This is hardly surprising.  Properties of the knowledge algorithms in
the framework of algorithmic knowledge immediately translate to
properties of the $X$ operator. To adapt an example from Halpern, Moses, and Vardi \citeyear{r:halpern94},  if a knowledge algorithm
is sound, that is, it says the agent explicitly knows $\phi$ in a state if and only if $\phi$ is true at that state, then $X\phi\rimp\phi$ is
valid in any structure using such a knowledge algorithm. What is
interesting in the context of deductive algorithmic knowledge is that
we can completely characterize the properties of $X$ by taking
advantage of the structure of the deductive systems. The remainder of
this section makes this statement precise.

As a first step, we introduce an axiomatization for reasoning about
deductive systems in general, independently of the actual deduction
rules of the system. This will form the basis of later
axiomatizations. First, we need axioms and inference rules capturing
propositional reasoning in the logic:
\begin{axiomlist}
\item[\axiom{Taut.}] All instances of propositional tautologies.
\item[\axiom{MP.}] From $\phi$ and $\phi\rimp\psi$ infer
$\psi$.
\end{axiomlist}
Axiom \axiom{Taut} can be replaced by an axiomatization of
propositional tautologies \cite{r:enderton72}. The following
well-known axioms and inference rules capture the properties of the
knowledge operator \cite{r:hintikka62}: 
\begin{axiomlist}
\item[\axiom{K1.}] $(K\phi\land K(\phi\rimp\psi))\rimp K\psi$.
\item[\axiom{K2.}] From $\phi$ infer $K\phi$.
\item[\axiom{K3.}] $K\phi\rimp\phi$.
\item[\axiom{K4.}] $K\phi\rimp K K\phi$.
\item[\axiom{K5.}] $\neg K\phi\rimp K\neg K\phi$.
\end{axiomlist}

We now turn to deductive algorithmic knowledge. Not surprisingly,
$X\phi$ does not satisfy many properties, because no assumptions were
made about the deductive systems. Deductive algorithmic
knowledge is interpreted with respect to the observations at the
current state, and two states are indistinguishable to an agent if the
same observations are made at both states; therefore, agents know whether or not
they explicitly know a fact. This is captured by the following axiom:
\begin{axiomlist}
\item[\axiom{X1.}] $X \phi \rimp K X \phi$.
\end{axiomlist}
In the presence of \axiom{K1}--\axiom{K5}, it is an easy exercise to
check that $\neg X\phi\rimp K\neg X\phi$ is provable from \axiom{X1}.
In addition, all observations are explicitly known. This fact is
expressed by the following axiom:
\begin{axiomlist}
\item[\axiom{X2.}] $\Ob(p)\rimp X\Ob(p)$.
\end{axiomlist}
Axiom \axiom{X2} just formalizes the following property of deduction as defined in
Section~\ref{s:deductivesystems}: for all terms $t$ of a
deductive system $D$, we have $t\vdash_D t$.  Finally, we need to
capture the fact that indistinguishable states have exactly
the same observations: 
\begin{axiomlist}
\item[\axiom{X3.}] $\Ob(p)\rimp K\Ob(p)$.
\end{axiomlist}
It is easy to see that the formula $\neg\Ob(p)\rimp K\neg\Ob(p)$ is
provable from \axiom{X3} in the presence of \axiom{K1}--\axiom{K5}.

\newcommand{\uMp}{M^\phi}
\newcommand{\Mc}{M^c}

Let $\AX$ consist of the axioms \axiom{Taut}, \axiom{MP},
\axiom{K1}--\axiom{K5}, and \axiom{X1}--\axiom{X3}. Without further
assumptions on the deductive systems under consideration, $\AX$
completely characterizes reasoning about deductive algorithmic knowledge.
\begin{theorem}\label{t:soundcomplete1}
The axiomatization $\AX$ is sound and complete for $\LKDprop(\Sigma)$
with respect to $\cM(\Sigma)$.
\end{theorem}
\begin{proof} 
See Appendix~\ref{a:section4}.
\end{proof}

If we want to reason about deductive algorithmic knowledge structures
equipped with a specific deductive system, we can say more. We can
essentially capture reasoning with respect to the specific deductive
system within our logic. The basic idea is to translate deduction
rules of the deductive system into formulas of $\LKDprop(\Sigma)$.  A
deduction rule of the form $t_1,\ldots,t_n\triangleright t$ in $D$ is
translated to a formula $(X t_1^R
\land\ldots\land X t_n^R)\rimp X t^R$, with the understanding that an
empty conjunction is just $\truep$. We define the formula $t^R$
corresponding to the term $t$ by induction on the structure of $t$:
$\OpTrue^R$ is $\truep$, $\OpFalse^R$ is $\falsep$, $(\OpNot(t))^R$ is
$\neg(t^R)$, $(\OpAnd(t_1,t_2))^R$ is $t_1^R\land t_2^R$,
$(\OpKnow(t))^R$ is $K(t^R)$, $(\OpDknow(t))^R$ is $X(t^R)$,
$(\OpOb(t))^R$ is $\Ob(t^R)$ (if $t\in T^g_\Sigma$), and $t^R$ is $t$
for all other terms $t$.  We view the result of the translation as an
axiom scheme, where the variables in $t_1,\ldots,t_n,t$ act as schema
metavariables, to be replaced by appropriate elements of the term
algebra.\footnote{One needs to be careful when defining this kind of
axiom schema formally. Intuitively, an axiom schema of the above form,
with metavariables appearing in terms, corresponds to the set of axioms
where each primitive proposition in the axiom is a ground substitution
instance of the appropriate term in the axiom schema.}  It is easy to
see that $(t^T)^R=t$ for all terms $t$. Furthermore, we do not
translate constructors in $\SKD$ that appear under constructors in
$\Sigma$ within a term. (Intuitively, these constructors will never
arise out of the translation of formulas given in
Section~\ref{s:deductive}.)  Let $\Axprop^D$ be the set of axioms
derived in this way for the $\SigmaKD$-deductive system $D$.

A simple argument shows that the axiomatization $\AX$ augmented with
axioms $\Axprop^D$ is not complete for $\cM_D(\Sigma)$, since there
are formulas of the form $X\psi$ that cannot be true in any structure
in $\cM_D(\Sigma)$, namely, $X\psi$ where $\psi^T$ is not derivable
from any set of observations using the deductive system $D$. Thus,
$\neg X\psi$ is valid for those $\psi$, but the axioms above clearly
cannot prove $\neg X\psi$. In other words, the axioms in $\Axprop^D$
capture deducibility in $\vdash_D$, but not nondeducibility. We can
however establish completeness with respect to a more general class of
structures, intuitively, those structures using a deductive system
containing \emph{at least} the deduction rules in $D$.  
Let $\cM_{D\subseteq}(\Sigma)$ be the class of all structures $M$ such
that there exists $D'$ with $D\subseteq D'$ and $M\in\cM_{D'}$.

\begin{theorem}\label{t:soundcomplete3}
The axiomatization $\AX$ augmented with axioms $\Axprop^D$ is sound
and complete for $\LKDprop(\Sigma)$ with respect to
$\cM_{D\subseteq}(\Sigma)$.  
\end{theorem}
\begin{proof}
See Appendix~\ref{a:section4}.
\end{proof}

If we are willing to restrict the formulas to consider, we can get
completeness with respect to $\cM_D(\Sigma)$. This follows directly
from the intuition that the axiomatization can prove all deductions in
$D$ but not the nondeductions. First, a few definitions: a
\emph{top-level} occurrence of a deductive algorithmic knowledge
subformula $X\psi$ in $\phi$ is an occurrence that does not occur in
the scope of an $X$ operator. An occurrence of a subformula $\psi$ is
said to be
\emph{positive} if it occurs in the scope of an even number of
negations. 

\begin{theorem}\label{t:restricted-completeness}
Let $\phi$ be a formula of $\LKDprop(\Sigma)$ in which every top-level
occurrence of a subformula $X\psi$ is positive; then $\phi$ is valid
in $\cM_D(\Sigma)$ if and only if $\phi$ is provable in the
axiomatization $\AX$ augmented with axioms $\Axprop^D$. 
\end{theorem}
\begin{proof}
See Appendix~\ref{a:section4}.
\end{proof}

In particular, Theorem~\ref{t:restricted-completeness} implies that a
formula of the form $X\phi$ is valid in $\cM_D(\Sigma)$ if and only if
$X\phi$ is provable in the axiomatization $\AX$ augmented with axioms
$\Axprop^D$.

\section{Decision Procedures}\label{s:complexity}

In this section, we study the decision problem for $\LKDprop(\Sigma)$, that
is, the problem of determining, for a given formula, whether it is
satisfiable. 
Again, we emphasize that we are considering only a single agent here;
allowing multiple agents changes the complexity, as we shall see in
Section~\ref{s:multiagent}. 
Since $\LKDprop(\Sigma)$ logic extends the logic of knowledge where the
knowledge operator is interpreted over an equivalence relation (in our
case, the relation $\sim$ saying that two states contain the same
observations), and since the complexity of the decision problem for
the latter is NP-complete \cite{r:ladner77}, the difficulty of
deciding satisfiability for $\LKDprop(\Sigma)$ is at least as hard.  We can use
the fact that there is a tight relationship between these logics to
get our complexity results.

We measure complexity in terms of the size of the formulas.
Define the size $\abs{t}$ of a term $t$ to be the
number of symbols required to write $t$, where each operation symbol
is counted as a single symbol. If $\Gamma$ is a finite set of terms, then
$\abs{\Gamma}$ is just the sum of the sizes of the terms in
$\Gamma$. Similarly, the size $\abs{\phi}$ of a formula is defined to be
the number of symbols required to write $\phi$, where again each
operation symbol is counted as a single symbol.

We can now state our complexity results. 
It turns out that adding a deductive algorithmic knowledge operator
to the logic of knowledge over an equivalence relation does not change
the complexity of the decision problem. 
Deciding satisfiability of a formula of $\LKDprop(\Sigma)$ is essentially
the same as deciding satisfiability of a formula in the logic of
knowledge over an equivalence relation, with the difference that to
account for deductive algorithmic knowledge, we need to construct a
deductive system with specific deduction rules that suffice to satisfy
the subformulas $X\phi$ appearing in the formula.

\begin{theorem}\label{t:decision1}
The problem of deciding whether a formula $\phi$ of $\LKDprop(\Sigma)$
is satisfiable  in $\cM(\Sigma)$ is NP-complete.
\end{theorem}
\begin{proof}
See Appendix~\ref{a:section5}.
\end{proof}

What happens if we fix a specific deductive system, and want to
establish whether a formula $\phi$ is satisfiable in a structure over
that deductive system? 
The difficulty of this problem depends intrinsically on the difficulty
of deciding whether a deduction $\Gamma\vdash_D t$ exists in $D$.
Since this problem may be arbitrarily difficult for certain deductive
systems $D$, reasoning in our logic can be arbitrarily difficult over
those deductive systems. 
The logic $\LKDprop(\Sigma)$ includes the propositional connectives,
which gives us an easy lower bound. 
\begin{theorem}\label{t:lowerbound}
For any given $\SigmaKD$-deductive system $D$, the problem of
deciding whether a formula $\phi$ of $\LKDprop(\Sigma)$ is satisfiable
in $\cM_D(\Sigma)$ is NP-hard.
\end{theorem}
\begin{proof}
See Appendix~\ref{a:section5}.
\end{proof}
On the other hand, if the deductive system is decidable in
nondeterministic polynomial time (i.e., if the problem of deciding
whether a deduction $\Gamma\vdash_D t$ exists in $D$ can be solved by
a nondeterministic Turing machine in time polynomial in 
$\abs{\Gamma}$ and $\abs{t}$), then the decision problem for
$\LKDprop$ remains relatively easy, at least with respect to models of
a reasonable size. More precisely, let $\cM^n_D(\Sigma)$ be the class of 
all deductive algorithmic knowledge structures using deductive system
$D$, where the number of observations at every state is at most $n$. 
\begin{theorem}\label{t:decision} 
For any given $\SigmaKD$-deductive system $D$ that is decidable in
nondeterministic polynomial time and for any polynomial $P(x)$, the
problem of deciding 
whether a formula $\phi$ of $\LKDprop(\Sigma)$ is satisfiable in
$\cM^{P(\abs{\phi})}_D(\Sigma)$ is NP-complete. 
\end{theorem}
\begin{proof}
See Appendix~\ref{a:section5}.
\end{proof}

There is a class of deductive systems that can be efficiently decided
(i.e., in polynomial time)
and thus by Theorem~\ref{t:decision} lead to a reasonable complexity
for $\LKDprop(\Sigma)$ interpreted over those systems.  Call a
deduction local in a deductive system $D$ if every proper subterm of a
term in the deduction is either a proper subterm of $t$, a proper
subterm of a member of $\Gamma$, or appears as a subterm of a
deduction rule in $D$. For any deductive system $D$, whether a local
deduction of $t$ from $\Gamma$ exists can be decided in time
polynomial in $\abs{\Gamma}$ and $\abs{t}$. A deductive system $D$ is
\emph{local} if whenever $\Gamma\vdash_D t$ there exists a local
deduction of $t$ from $\Gamma$ \cite{r:mcallester93}. Thus, if $D$ is
a local deductive system, the existence of a deduction ensures the
existence of a local deduction, and consequently the deduction
relation $\vdash_D$ is polynomial-time decidable. The deductive system
in Example~\ref{x:dy} is local, while adding the deduction rules in
Example~\ref{x:bool} to any local deductive system yields a local
deductive system.

\begin{theorem}\label{c:decision} 
For any local $\SigmaKD$-deductive system $D$ and for any polynomial
$P(x)$, the problem of 
deciding whether a formula $\phi$ of $\LKDprop(\Sigma)$ is satisfiable
in $\cM^{P(\abs{\phi})}_D(\Sigma)$ is NP-complete.
\end{theorem}
\begin{proof}
Immediate from the property of local deductive system, and from
Theorem~\ref{t:decision}. 
\end{proof}

\section{Reasoning about Multiple Agents}\label{s:multiagent}

The framework we have described extends to multiple agents in a
straightforward way. This extension is similar to the extension of
modal logics of knowledge to multiple agents \cite{r:fagin95}.  The
only addition is that we need to equip every agent with a deductive
system. 

Suppose a group of agents, named $1,\dots,n$ for simplicity. 
We define the logic $\LKDnprop(\Sigma)$ as we did
$\LKDprop(\Sigma)$, except that the operators $K_i$, $X_i$, and
$\Ob_i$ are indexed by an agent. A priori, there is no difficulty in
giving a semantics to this logic as we have done in
Section~\ref{s:deductive}. Unfortunately, this does not let an agent
explicitly reason about another agent's knowledge. In order to do
this, we need to modify and extend the framework. As before, we
consider deductive systems over a signature $\Sigma$ extended with a
set $\SKDn$ of constructors given by
$\SKDn=\{\OpTrue, \OpFalse, \OpNot,
\OpAnd\}\cup\bigcup_{i=1}^{n}\{\OpOb_i,\OpKnow_i,\OpDknow_i\}$, where
$\OpTrue,\OpFalse$ have arity $0$,
$\OpOb_i,\OpNot,\OpKnow_i,\OpDknow_i$ have arity 1, and $\OpAnd$ has
arity $2$. 

A \emph{deductive algorithmic knowledge structure with $n$ agents} is
a tuple $M=(S,\pi,D_1,\dots,D_n)$, where $S$ is a set of states, $\pi$
is an interpretation for the primitive propositions, and $D_i$ is a
$\SigmaKDn$-deductive system. Every state $s$ in $S$ is of the form
$(e,O_1,\dots,O_n)$, where $e$ captures the general state of the
system, and $O_i$ is a 
finite set of observations from $T^g_\Sigma$, representing the
observations that agent $i$ has made at that state. The interpretation
$\pi$ associates with every state the set of primitive propositions
true at that state, so that for all primitive proposition $p\in
T^g_\Sigma$, we have $\pi(s)(p)\in\{\true,\false\}$. 

Let $\cM_n(\Sigma)$ be the set of all deductive algorithmic knowledge
structures using $\SigmaKDn$-deductive systems for each agent. For fixed
$\SigmaKDn$-deductive systems $D_1,\dots,D_n$, let
$\cM_{D_1,\dots,D_n}(\Sigma)$ be the set of all deductive algorithmic
knowledge structures for $n$ agents with deductive systems
$D_1,\dots,D_n$ (i.e., agent $i$ uses deductive system
$D_i$).

The remaining definitions generalize in a similar way. We define, for
each agent, a relation on the states that captures the states that the
agent cannot distinguish, based on his observations. More precisely,
let $s\sim_i s'$ if and only $s=(e,O_1,\dots,O_n)$ and
$s'=(e',O'_1,\dots,O'_n)$, for some $e,e'$ and sets of
observations $O_1,\dots,O_n,O'_1,\dots,O'_n$ with
$O_i=O'_i$. Again, $\sim_i$ is an equivalence relation on the
states.

The translation of a formula $\phi$ into a term $\phi^T$ of the
deductive system now takes into account the name of the agents. As
expected,  $p^T$ is $p$ for any primitive proposition $p$, 
$(\neg\phi)^T$ is $\OpNot(\phi^T)$, $(\phi\land\psi)^T$ is
$\OpAnd(\phi^T,\psi^T)$, $(K_i\phi)^T$ is $\op{know}_i(\phi^T)$, 
$(X_i\phi)^T$ is $\OpDknow_i(\phi^T)$, and $(\Ob_i(p))^T$ is
$\OpOb_i(p)$. 

The semantics is just like that of Section~\ref{s:deductive}, except
with the following rules for $K_i\phi$, $X_i\phi$, and $\Ob_i(p)$:
\begin{itemize}
\item[] $(M,s)\sat K_i\phi$ if $(M,s')\sat\phi$ for all $s'\sim_i s$
\item[] $(M,s)\sat X_i\phi$ if $s=(e,O_1,\dots,O_n)$ and
$\{\OpOb_i(p)\mid p\in O_i\}\vdash_{D_i} \phi^T$
\item[] $(M,s)\sat \Ob_i(p)$ if $s=(e,O_1,\dots,O_n)$ and
$p\in O_i$. 
\end{itemize}

\begin{example}\label{x:new2}
Kaplan and Schubert \citeyear{r:kaplan00} study a phenomenon they call 
\emph{simulative inference} where, roughly speaking, an agent can
reconstruct the reasoning of another agent. We can capture this phenomenon by
making suitable assumptions on an agent's deductive system. (Kaplan
and Schubert work in a different setting---they assume that the
inference engine is explicitly told formulas, and thus 
work in a setting similar to that of belief revision \cite{agm:85}.)
Say that a deductive system $D_i$ for agent $i$ \emph{permits
simulative inference of agent $j$ with $D_j$} if $D_i$ contains a rule
$\OpOb_j(t)\triangleright\OpDknow_j(\OpOb_j(t))$, and for every rule
$t_1,\dots,t_k\triangleright t$ of $D_j$, there is a corresponding
rule
$\OpDknow_j(t_1),\dots,\OpDknow_j(t_k)\triangleright\OpDknow_j(t)$ in
$D_i$. It is then easy to check that if we have $(M,s)\sat X_j\phi$
for some state $s=(e,O_1,\dots,O_n)$ with
$\{p_1,\dots,p_k\}\subseteq O_j$, and $(M,s)\sat X_i\Ob_j(p_1)\land
\dots\land X_i\Ob_j(p_k)$, then $(M,s)\sat X_i X_j
\phi$. Note that this derivation assumes that the agent $i$ can
explicitly determine that agent $j$ has observed
$p_1,\dots, p_k$. 
\end{example}

As far as axiomatizations are concerned, we can essentially lift the
results of Section~\ref{s:axiomatizations}. It suffices to consider an
axiomatization where \axiom{K1}--\axiom{K5} now refer to $K_i$ rather
than just $K$. For instance, \axiom{K1} becomes $K_i\phi\land
K_i(\phi\rimp\psi)\rimp K_i\psi$, for every agent $i$. In a similar
way, the axiom \axiom{X1} simply becomes $X_i\phi\rimp K_i
X_i\phi$. For \axiom{X2} and \axiom{X3}, we need to further restrict
the observations to be those of the agent under consideration:
$\Ob_i(p)\rimp X_i\Ob_i(p)$, and $\Ob_i(p)\rimp K_i\Ob_i(p)$. Let
$\AX_n$ be the resulting axiomatization.

\begin{theorem}\label{t:another-axiomatization}
The axiomatization $\AX_n$ is sound and complete for
$\LKDnprop(\Sigma)$ with respect to $\cM_n(\Sigma)$.  
\end{theorem}
\begin{proof}
See Appendix~\ref{a:section6}.
\end{proof}

As in the single agent case, we can capture the reasoning with respect
to specific deductive systems (one per agent) within our logic. Again,
we translate deduction rules of the deductive systems into formulas of
$\LKDnprop(\Sigma)$.
Consider the deductive system $D_i$ for agent $i$. A deduction rule of
the form $t_1,\ldots,t_n\triangleright t$ in $D_i$ is translated to a
formula $(X_i t_1^R
\land\ldots\land X_i t_n^R)\rimp X_i t^R$. We define the formula $t^R$
corresponding to the term $t$ by induction on the structure of $t$:
$\OpTrue^R$ is $\truep$, $\OpFalse^R$ is $\falsep$, $(\OpNot(t))^R$ is
$\neg(t^R)$, $(\OpAnd(t_1,t_2))^R$ is $t_1^R\land t_2^R$,
$(\OpKnow_i(t))^R$ is $K_i(t^R)$, $(\OpDknow_i(t))^R$ is $X_i(t^R)$,
$(\OpOb_i(t))^R$ is $\Ob_i(t^R)$ (if $t\in T^g_\Sigma$), and $t^R$ is
$t$ for all other terms $t$. (As in Section~\ref{s:axiomatizations},
such a translation yields an axiom schema, where we view the variables
in $t_1,\ldots,t_n,t$ as schema metavariables, to be replaced by
appropriate elements of the term algebra.)  Let $\Axprop^{D_i}_n$ be
the set of axioms derived in this way for the deductive system $D_i$
of agent $i$.

As in the single agent case, we cannot capture exactly the reasoning
in structures where agent $i$ is using deductive system $D_i$, since
we cannot capture nondeducibility within the logic. Therefore,
completeness is established with respect to a larger class of
structures. 
Let $\cM_{D_1,\dots,D_n\subseteq}(\Sigma)$ be the class of all
structures $M$ such that there exists $D'_1,\dots,D'_n$ with
$D_1\subseteq D'_1, \dots, D_n\subseteq D'_n$ and
$M\in\cM_{D'_1,\dots,D'_n}$. 

\begin{theorem}\label{t:another-axiomatization2}
The axiomatization
$\AX_n$ augmented with axioms $\Axprop^{D_1}_n,\dots,\Axprop^{D_n}_n$
is sound and complete for $\LKDnprop(\Sigma)$ with respect to
$\cM_{D_1,\dots,D_n\subseteq}(\Sigma)$.   
\end{theorem}
\begin{proof}
See Appendix~\ref{a:section6}.
\end{proof}

The complexity of the decision problem in the case of multiple agents
reflects the complexity of the decision problem for the modal logic of
knowledge with multiple agents.  The logic $\LKDnprop(\Sigma)$ extends
the logic of knowledge over equivalence relations for $n$ agents, and
it is known that the decision problem for that logic is
PSPACE-complete \cite{r:halpern92}. As in the single agent case,
adding deductive algorithmic knowledge does not affect the complexity
of the decision problem with respect to arbitrary deductive systems.

\begin{theorem}\label{t:complexity-multiagent}
If $n\ge 2$, the problem of deciding whether a formula $\phi$ of
$\LKDprop_n(\Sigma)$ is satisfiable  in $\cM_n(\Sigma)$ is PSPACE-complete.
\end{theorem}
\begin{proof}
See Appendix~\ref{a:section6}.
\end{proof}

There is no clear candidate for an equivalent of
Theorem~\ref{t:decision} in the multiple agents context. Assuming
every agent uses a tractable deductive system yields an easy EXPTIME
upper bound on the decision problem for $\LKDprop_n(\Sigma)$, while the 
best lower bound we obtain is the same as the one in
Theorem~\ref{t:complexity-multiagent}, that is, the problem is
PSPACE-hard.

\section{Conclusion}\label{s:conclusion}

We have described in this paper an approach to combining implicit
knowledge interpreted over possible worlds with a notion of explicit
knowledge based on a deductive system that allows agents to derive
what they explicitly know. This additional structure, the agent's
deductive system, can be used to completely characterize the
properties of the explicit knowledge operator. More specifically, we
can derive sound and complete axiomatizations for the logic in a
uniform way. There are many approaches to modeling  explicit knowledge in the
literature, with many different philosophical intuitions and
interpretations. The model in this paper is based on a 
conception of explicit knowledge (viz., obtained from inference rules) that
goes back at least to Konolige \citeyear{r:konolige86}, and
aims at capturing a particularly computational interpretation of
explicit knowledge based on observations. 
We remark that our model is consistent with a number of
epistemological theories \cite{r:pollock99} that argue that all
knowledge is ultimately derived from observations. 

While the framework extends to multiple agents in a straightforward
way, there are many issues that still remain to be addressed at that
level. For instance, a natural question is what happens when we move
to more dynamic models, where the observations are taken over
time. There are interesting issues that arise, especially when we
assume that agents do not share a clock to synchronize their
observations. We hope to explore this extension in future work.

What is our framework good for? Because deductive algorithmic
knowledge is a special case of algorithmic
knowledge, any situation that can be modeled in our framework can also
be modeled in the original algorithmic knowledge framework.
The one advantage of our logic, however, is that it
admits sound and complete axiomatizations derivable directly from the
deductive systems. Therefore, we can devise useful proof
systems for interesting classes of applications, corresponding to
different deductive systems.

The framework in this paper sheds light on the epistemic content of
deductive systems, in that we provide a logic in which we can reason
about implicit knowledge and explicit knowledge derived from a
deductive system. An interesting question is whether this approach can
shed light on the epistemic content of probabilistic deductive
systems, of the kind found in the recent probabilistic deductive
database literature \cite{r:lukasiewicz99,r:lakshmanan01}. Presumably,
the ideas developed by Halpern and Pucella \citeyear{r:halpern05c} may be applicable in this
setting.

\paragraph*{Acknowledgments.} Thanks to Hubie Chen and Vicky Weissman for
comments on previous drafts of this paper. Special thanks to Joseph Halpern for 
carefully reading a version of this paper and pointing out problems with
the original statements of Theorems~\ref{t:soundcomplete3} and
\ref{t:another-axiomatization2}, and for suggesting
Theorem~\ref{t:restricted-completeness}.  Anonymous referees also
did a wonderful job of suggesting improvements. This work was
supported in part by NSF under grant CTC-0208535, by ONR under grant
N00014-02-1-0455, by the DoD Multidisciplinary University Research
Initiative (MURI) program administered by the ONR under grant
N00014-01-1-0795, and by AFOSR under grant F49620-02-1-0101.

\appendix

\section{Proofs for Section~\ref{s:axiomatizations}}
\label{a:section4}

The proof of soundness and completeness in
Section~\ref{s:axiomatizations} rely on fairly standard canonical
model constructions \cite{r:hughes96}. 
We review the required notions here, for completeness.
A canonical model
is a model whose states are consistent sets of formulas, with the
property that valid formulas of the logic are true at states of the
canonical model.  Recall that a formula $\phi$ is consistent (with
respect to an axiomatization $\AX$) if $\neg\phi$ is not provable from
$\AX$. A finite set of formulas $\{\phi_1,\dots,\phi_n\}$ is
consistent if and only if $\phi_1\land\dots\land\phi_n$ is
consistent. An infinite set $F$ of formulas is consistent if and only
if every finite subset of $F$ is consistent. A set of formulas is
maximally consistent if it is not properly contained in any other
consistent set of formulas. We assume some familiarity with properties
of maximally consistent sets of formulas. Such properties include, for
$V$ a maximally consistent set of formulas: for all $\phi$, exactly
one of $\phi$ or $\neg\phi$ is in $V$; $\phi$ and $\psi$ are both in
$V$ if and only if $\phi\land\psi$ is in $V$; if $\phi$ and
$\phi\rimp\psi$ are in $V$, then $\psi$ is in $V$; every provable
formula is in $V$ (in particular, every axiom of $\AX$ is in $V$). 

Some definitions will be useful.  First, given a set $V$ of formulas,
let $V/K=\{\phi\mid K\phi\in V\}$.  Let $\C$ be the set of all maximal
consistent sets of formulas of $\LKDprop(\Sigma)$.  For $V\in\C$, let
$V/\Ob=\{p\in T^g_\Sigma\mid \Ob(p)\in V\}$.  Define the relation
$\approx$ over $\C$ by taking $V\approx U$ if and only if
$V/K\subseteq U$.

\begin{lemma}\label{l:approx-props}
\begin{enumerate}
\item $\approx$ is an equivalence relation on $\C$.
\item If $V\approx U$, then $V/\Ob=U/\Ob$. 
\item For all $\psi$ in
$\LKDprop(\Sigma)$, if $V\approx U$, then $X\psi\in V$ if and only if
$X\psi\in U$.
\end{enumerate}
\end{lemma}
\begin{proof}
(1) The proof that $\approx$ is an equivalence relation is completely
  standard, and relies on the axioms \axiom{K1}--\axiom{K5}
  \cite{r:hughes96}. 

(2) Let $p\in V/\Ob\subseteq V$. Since $V$ is maximally consistent,
all instances of \axiom{X3} are in $V$, and thus $\Ob(p)\rimp K\Ob(p)$
is in $V$, so by \axiom{MP}, $K\Ob(p)\in V$, and thus $\Ob(p)\in
V/K\subseteq U$. Therefore, $\Ob(p)\in U$, and
$V/\Ob \subseteq U/\Ob$. Since $\approx$ is an equivalence relation,
$V\approx U$ implies that $U\approx V$, and by the same argument, we
get $U/\Ob\subseteq V/\Ob$.

(3)  This follows easily from \axiom{X1}. Assume $X\psi\in
V$. Then $KX\psi\in V$ by \axiom{X1}, and thus $X\psi\in V/K$, and
since $V\approx U$, $X\psi\in U$. The converse direction follows from
the fact that $\approx$ is symmetric.
\end{proof}

\begin{oldtheorem}{t:soundcomplete1}
\begin{theorem}
The axiomatization $\AX$ is sound and complete for $\LKDprop(\Sigma)$ with
respect to $\cM(\Sigma)$. 
\end{theorem}
\end{oldtheorem}
\begin{proof} 
Proving soundness is straightforward.  For completeness, we prove the
equivalent statement that if $\phi$ is consistent (i.e., $\neg\phi$ is
not provable from the axiomatization $\AX$), then $\phi$ is satisfiable in
some structure in $\cM(\Sigma)$. The satisfying structure will be
constructed from the set of maximally consistent formulas.

Let $\phi$ be a consistent formula of $\LKDprop(\Sigma)$,
and let $\mathit{Sub}(\phi)$ be the set of subformulas of $\phi$
(including $\phi$ itself).
Since $\phi$ 
is consistent, there is a set $V^\phi\in 
\C$ with $\phi\in V^\phi$
with $\abs{V^\phi/\Ob}<\infty$: first construct a set $V$ starting with
$\phi$, adding $\Ob(p)$ for every observation $\Ob(p)$ appearing in
$\phi$ if $\Ob(p)\land\phi$ is consistent, and adding $\neg\Ob(p)$ for every
observation $\Ob(p)$ either not appearing in $\phi$ or inconsistent with
$\phi$; it easy to establish that $V$ is consistent, so $V$ is
extensible to a maximally consistent set $V^\phi$ with
$\abs{V^\phi/\Ob}<\infty$.

Let $O^\phi=V^\phi/\Ob$.  Let $[V^\phi]_\approx$ be the
$\approx$-equivalence class that contains $V^\phi$. We will use the
sets of formulas $[V^\phi]_\approx$ as states of our structure. More
specifically, define the deductive algorithmic knowledge structure
$M^\phi=(S^\phi,\pi^\phi,D^\phi)$ by taking:
\begin{align*}
 & S^\phi  = \{ (s_V,O^\phi) \mid V\in[V^\phi]_\approx\}\\
 & \pi^\phi((s_V,O^\phi))(p)  = \begin{cases}
                  \true & \text{if $p\in V$}\\
                  \false & \text{if $p\not\in V$}
                  \end{cases}\\
 & D^\phi  = \{ (\emptyset,\psi^T) \mid X\psi\in\mathit{Sub}(\phi),
X\psi\in V^\phi\}. 
\end{align*}
To simplify the discussion, and because $O^\phi$ is fixed in
$M^\phi$, we refer to the state $(s_V,O^\phi)$ as simply $s_V$; for
instance, we freely write $s_V\in S^\phi$.  We can check that $D^\phi$
defines a $\SigmaKD$-deductive system.  Decidability of $D^\phi$ holds
trivially, since $D^\phi$ contains finitely many deduction rules, as
$\mathit{Sub}(\phi)$ is finite.
Thus, $M^\phi$ is a deductive algorithmic knowledge structure.

We now show that for all $s_V\in S^\phi$
and all subformulas $\psi\in\mathit{Sub}(\phi)$,
we have 
$(\uMp,s_V)\sat\psi$
if and only if $\psi\in V$, by induction on the structure of formulas.
The cases for $\truep$, $\falsep$, primitive propositions,
conjunction, and negation are straightforward, using maximal
consistency of $V$.

For $\Ob(p)$, first assume that $(M^\phi,s_V)\sat\Ob(p)$. This means
that $p\in O^\phi$, so $p\in V^\phi/\Ob$, and since
$V^\phi/\Ob=V/\Ob$, $p\in V/\Ob$, and thus $\Ob(p)\in
V$. Conversely, if $\Ob(p)\in V$, then $p\in V/\Ob=V^\phi/\Ob$, so
that $p\in O^\phi$, and thus $(M^\phi,s_V)\sat\Ob(p)$.

Now, consider a deductive algorithmic knowledge formula
$X\psi$. First, assume that we have $(\uMp,s_V)\sat X\psi$.  By
definition, $\{\OpOb(p)\mid p\in O^\phi\}\vdash_{D^\phi}\psi^T$. If
$\psi$ is $\Ob(p)$ for some $p\in O^\phi$, then
$(M^\phi,s_V)\sat\Ob(p)$, and thus $\Ob(p)\in V$ by the induction
hypothesis; by \axiom{X2}, $X\Ob(p)\in V$. Otherwise, by construction
of $D^\phi$, there must exist a rule $\triangleright \psi^T$ in
$D^\phi$. In other words, $X\psi\in V^\phi$. Since $V\approx V^\phi$
by choice of $S^\phi$, we get $X\psi\in V$, following
Lemma~\ref{l:approx-props}.  Conversely, assume that $X\psi\in V$. By
definition of $D^\phi$, $(\emptyset,\psi^T)\in D^\phi$, and thus
$\{\OpOb(p)\mid p\in O^\phi\}\vdash_{D^\phi}\psi^T$, meaning that
$(\uMp,s_V)\sat X\psi$.

For a knowledge formula $K\psi$, the result follows from
essentially the same proof as that of Halpern and Moses
\citeyear{r:halpern92}.  We give it here for completeness. First,
assume $(\uMp,s_V)\sat K\psi$. It 
follows that $(V/K)\cup\{\neg\psi\}$ is not consistent. (Otherwise, it
would be contained in some maximal consistent set $U$ in $\C$, and by
construction, we would have $V/K\subseteq U$, and thus $V\approx U$,
and hence $s_V\sim s_U$; but since we have $\neg\psi\in U$, we have
$\psi\not\in U$, and by the induction hypothesis,
$(\uMp,s_U)\not\sat\psi$, contradicting $(\uMp,s_V)\sat K\psi$.) Since
$(V/K)\cup\{\neg\psi\}$ is not consistent, there must be some finite
subset $\{\phi_1,\dots,\phi_k,\neg\psi\}$ which is not consistent. By
propositional reasoning, we can derive that
$\phi_1\rimp(\phi_2\rimp(\dots\rimp (\phi_k\rimp\psi)\dots))$ is
provable, and thus $K(\phi_1\rimp(\phi_2\rimp(\dots\rimp
(\phi_k\rimp\psi)\dots)))$ is provable by \axiom{K2}. It is
straightforward to derive from this by induction, propositional
reasoning, and \axiom{K1}, that
$K\phi_1\rimp(K\phi_2\rimp(\dots\rimp(K\phi_k\rimp K\psi)\dots))$ is
provable. Thus, $K\phi_1\rimp(K\phi_2\rimp(\dots\rimp(K\phi_k\rimp
K\psi)\dots))\in V$. Because $\phi_1,\dots,\phi_k\in V/K$, we have
$K\phi_1,\dots,K\phi_k\in V$, and by repeated applications of
\axiom{MP}, we have $K\psi\in V$, as desired.  Conversely, if we
assume $K\psi\in V$, then $\psi\in V/K$. Let $s_U$ be an arbitrary
state of $S^\phi$. By construction of $\uMp$, $V\approx U$ and thus
$V/K\subseteq U$. Therefore, we have $\psi\in U$, and by the induction
hypothesis, $(\uMp,s_U)\sat\psi$. Since $s_U$ was arbitrary, and since
$s_U\sim s_V$ (immediate by the definition of $S^\phi$), this means
that $(\uMp,s_V)\sat K\psi$.

Completeness of $\AX$ now follows immediately. Since $\phi\in V^\phi$
and $\phi\in\mathit{Sub}(\phi)$,
we have $(\uMp,s_{V^\phi})\sat\phi$, and thus $\phi$ is satisfiable. 
\end{proof}

\begin{oldtheorem}{t:soundcomplete3}
\begin{theorem}
The axiomatization $\AX$ augmented with axioms $\Axprop^D$ is sound
and complete for $\LKDprop(\Sigma)$ with respect to
$\cM_{D\subseteq}(\Sigma)$.  
\end{theorem}
\end{oldtheorem}
\begin{proof}
Soundness is again straightforward. 
For completeness, we prove
the equivalent statement that if $\phi$ is consistent (i.e., if
$\neg\phi$ is not provable from the axiomatization $\AX$ augmented
with the axioms $\Axprop^D$) then 
$\phi$ is satisfiable in some structure in $\cM_D(\Sigma)$. The procedure
is exactly the one that is used to prove
Theorem~\ref{t:soundcomplete1}, except with a different deductive
system $D^\phi$.

We simply indicate where the proof differs from that of
Theorem~\ref{t:soundcomplete1}, and let the reader fill in the
details. We construct, for a given $\phi$, a deductive algorithmic
knowledge structure $M^\phi=(S^\phi,\pi^\phi,D^\phi)$, where
$S^\phi$ and $\pi^\phi$ are constructed as in
Theorem~\ref{t:soundcomplete1}, and $D^\phi$ is given by 
\[ D\cup\{ (\emptyset,\psi^T) \mid X\psi\in\mathit{Sub}(\phi),
X\psi\in V^\phi\}.\] We can check that $M^\phi$ is a deductive
algorithmic knowledge structure in $\cM_{D\subseteq}(\Sigma)$.  The
deductive system $D^\phi$ has the following interesting property: if
$\{\OpOb(p)\mid p\in O^\phi\}\vdash_D\psi^T$, then there is a rule
$\triangleright\psi^T$ in $D^\phi$. In other words, every term
$\psi^T$ derivable from the rules in $D$ is derivable directly with a
single rule in $D^\phi$. Here is the proof of this property. Assume
$\{\OpOb(p)\mid p\in O^\phi\}\vdash_D\psi^T$. Clearly, it is
sufficient to show that $X\psi\in V^\phi$.  If $\psi$ is $\Ob(p)$ for
some $p\in O^\phi$, then $\Ob(p)\in V^\phi$, so by \axiom{X2},
$X\Ob(p)\in V^\phi$. Otherwise, there must exist a deduction
$t_1,\dots,t_m$ in $D$ such that $t_m=\psi^T$ is a conclusion of the
deduction. We show by induction on the length of the deduction that
for every $i$, $X(t_i)^R\in V^\phi$. For the base case $i=1$, we have
two cases. If $t_1$ is $\OpOb(t)$ for some $t\in O^\phi$, then $(t_1)^R=\Ob(t)$, 
and $X \Ob(t)\in V^\phi$ follows from \axiom{X2} and the fact that
$t\in O^\phi$ implies that $\Ob(t)\in V^\phi$.  
If $t_1$ follows from the
application of a deduction rule in $D$, with no antecedents, then by
construction, there is an instance of this rule in $V^\phi$, of the
form $\truep\rimp X(t_1)^R$, and thus $X(t_1)^R\in V^\phi$. For the
inductive case $i>1$, again, there are two cases. If $t_i$ is
$\OpOb(t)$ for some $t\in O^\phi$, then the result $X(t_1)^R$
follows as in the base case.  Otherwise, there is a
rule $t_1',\dots,t_k'\triangleright t'$ in $D$ such that for some
ground substitution $\rho$ such that $\rho(t')=t_i$ and for all $j\in
1..k$, $\rho(t_j')$
appears in the deduction before term $t_i$. By construction, there is
an instance of $X(t_1')^R\land\dots\land X(t_k')^R\rimp X(t')^R$ in
$V^\phi$, and by induction hypothesis, we have $X(t_{i_j}')^R\in V^\phi$
for each $i_j<i$. Thus, by \axiom{MP}, we have $X(t_i)^R\in
V^\phi$. Since $\psi^T=t_m$, the last element of the deduction, we get
that $X(\psi^T)^R = X\psi$ is in $V^\phi$, as desired.

The rest of the proof follows as before. 
\end{proof}

The following lemma is useful for proving
Theorem~\ref{t:restricted-completeness}. Roughly, it says that any
formula can be written as an equivalent formula where all instances
of the negation operator occurs before a primitive proposition, an
observation formula, or an explicit knowledge formula. 
To simplify the presentation of the lemma, we write the resulting
formula using operator $\lor$, as well as operator $L\phi$, which is
just an abbreviation  for $\neg K\neg \phi$.  
\begin{lemma}\label{l:rewrite}
Every formula $\phi$ of $\LKDprop(\Sigma)$ is logically equivalent to
a formula $\overline{\phi}$ written using $p$, $\neg p$, $\Ob(p)$,
$\neg\Ob(p)$ (for primitive propositions $p$) and $X\phi$ and $\neg
X\phi$ (for $\phi$ in $\LKDprop(\Sigma)$), and the operators $\land$,
$\lor$, $K$, and $L$. Moreover, if every top-level occurrence of a
subformula $X\psi$ is positive in $\phi$, then no occurrence of
$X\psi$ in $\overline{\phi}$ is in the scope of a negation.
\end{lemma}
\begin{proof}
The first part of the lemma follows directly from the following
laws, which permit one to move negations into a formula as far down as
they go:
\begin{align*}
& \neg(\phi\land\psi) \riff (\neg\phi)\lor(\neg\psi)\\
& \neg(\phi\lor\psi) \riff (\neg\phi)\land(\neg\psi)\\
& \neg(K\phi) \riff L(\neg\phi)\\
& \neg(\neg\phi) \riff \phi.
\end{align*}
The second part of the lemma follows by observing that the laws above
preserve the evenness or oddness of the number of negations
under the scope of which each subformula $X\psi$ appears.
\end{proof}

\begin{oldtheorem}{t:restricted-completeness}
\begin{theorem}
Let $\phi$ be a formula of $\LKDprop(\Sigma)$ in which every top-level
occurrence of a subformula $X\psi$ is positive; then $\phi$ is valid
in $\cM_D(\Sigma)$ if and only if $\phi$ is provable in the
axiomatization $\AX$ augmented with axioms $\Axprop^D$. 
\end{theorem}
\end{oldtheorem}
\begin{proof}
Let $\phi$ be of the required form. By Lemma~\ref{l:rewrite},
$\overline{\phi}$ is also of the required form, and moreover is such
that negations only appear in front of $p$, $\Ob(p)$, or
$X\psi$. Since each occurrence is $X\psi$ in $\overline{\phi}$ is
positive, this means that every $X\psi$ appears unnegated in
$\overline{\phi}$. Let $M=(S,\pi,D)$ be an arbitrary model in
$\cM_D(\Sigma)$, and let $M_{D'}=(S,\pi,D')$ be the corresponding
model for $D'\supseteq D$. We claim that for all $s\in S$,
$(M,s)\sat\overline{\phi}$ if and only if
$(M_{D'},s)\sat\overline{\phi}$. This is easily established by induction
on the structure of $\overline{\phi}$. (The key point is that we do
not need to consider the case $\neg X\psi$, which is guaranteed not to
occur in $\overline{\phi}$.) Since $s$ was arbitrary, we have
$M\sat\overline{\phi}$ if and only if
$M_{D'}\sat\overline{\phi}$. Now, assume $\phi$ is valid in
$\cM_D(\Sigma)$. Let $M_{D'}=(S,\pi,D')$ be an arbitrary model in
$\cM_{D\subseteq}(\Sigma)$, and let $M=(S,\pi,D)$. Since
$\overline{\phi}$ is valid in $\cM_D(\Sigma)$,
$M\sat\overline{\phi}$; by our result above,
$M_{D'}\sat\overline{\phi}$. Since $M_{D'}$ was arbitrary, we have
$\overline{\phi}$ is valid in $\cM_{D\subseteq}(\Sigma)$. 

Conversely, assume $\phi$ is provable in the axiomatization $\AX$
augmented with axioms $\Axprop^D$. By Theorem~\ref{t:soundcomplete3},
$\phi$ is valid in $\cM_{D\subseteq}(\Sigma)$. Since
$\cM_D(\Sigma)\subseteq\cM_{D\subseteq}(\Sigma)$, then $\phi$ is
certainly valid in $\cM_D$. 
\end{proof}

\section{Proofs for Section~\ref{s:complexity}}
\label{a:section5}

We assume the terminology and notation of \cite{r:fagin95} for the
modal logic of knowledge over an arbitrary equivalence relation; we
call this logic $\LK$, and let $f,g$ range over formulas of
$\LK$. 
(This notation will let us distinguish $\LK$ formulas from $\LKDprop(\Sigma)$
formulas when they occur in the same statements.)
The logic $\LK$ is interpreted over $\LK$-structures, that is, Kripke
structures $(S,\cK,\pi)$ with an arbitrary equivalence relation $\cK$
over the states that is used to interpret the modal operator; the
satisfaction relation is written $(M,s)\satS f$, and is defined just
like $\sat$, but using the equivalence relation $\cK$ over the states
rather than the $\sim$ relation, and without the $X$ and $\Ob$
operators. We write $\cK(s)$ for $\{s'\mid (s,s')\in\cK\}$.  The
following small model result for $\LK$, due to Ladner
\citeyear{r:ladner77}, is central to most of the proofs in this
section.
\begin{theorem}\label{p:ladner}
{\rm\cite{r:ladner77}} Given $f$ an $\LK$ formula, if $f$ is
satisfiable, then $f$ is satisfiable in an $\LK$-structure
$M=(S,\cK,\pi)$ where $\abs{S}\le \abs{f}$, and $\cK$ is the universal
relation, that is, $\cK=S\times S$. 
\end{theorem}

Our result will follow by relating the complexity of
$\LKDprop(\Sigma)$ to the complexity of $\LK$. 
We can easily reduce the decision problem for $\LK$ to our
logic, by simply ignoring the $X\phi$ formulas. Consider the following
construction.  Let $f$ be a formula of $\LK$. Let $p_1,\ldots,p_k$
be the primitive propositions appearing in $f$. We first come up
with an encoding of these primitive propositions into the language of
$\Sigma$. For example, we can take $p_1$ to be $\OpTrue$, $p_2$ to be
$\OpNot(\OpTrue)$, $p_3$ to be $\OpNot(\OpNot(\OpTrue))$, and so
forth. Let $t_p$ be the term encoding the primitive proposition
$p$. 
Let $\hat{f}$ be the formula obtained by replacing every 
instance of a primitive proposition $p$ in $f$ by $t_p$. Note that
$\abs{\hat{f}}$ is polynomial in $\abs{f}$, 
and that $\hat{f}$ contains no instance of the $X$ operator.
\begin{lemma}\label{l:dedx1}
Given $f$ an $\LK$ formula, and given $D$ an arbitrary KD deductive
system over $\Sigma$, the following are equivalent:
\begin{enumerate}
\item $f$ is satisfiable in an $\LK$-structure;
\item $\hat{f}$ is satisfiable in $\cM_D(\Sigma)$;
\item $\hat{f}$ is satisfiable in $\cM(\Sigma)$.
\end{enumerate}
\end{lemma}
\begin{proof}
$(1)\rimp(2)$: Assume $f$ is satisfiable in an $\LK$-structure.  By
Theorem~\ref{p:ladner}, we know that there exists an $\LK$-structure
$M=(S,\cK,\pi)$ where $\cK$ is an equivalence relation on $S$ and
$(M,s)\satS f$ for some $s\in S$.\footnote{While
Theorem~\ref{p:ladner} says that the equivalence relation $\cK$ can be
taken to be universal, we will not take advantage of this in this
proof or the proof of Lemma~\ref{l:dedx2}. This is in order to
simplify the generalization of these proofs to the multiple agents
case (Theorem~\ref{t:complexity-multiagent}).}  Let $\{ [s]_{\cK} \mid
s\in S\}$ be the set of equivalence classes of $\cK$, of which there
are at most $\abs{f}$. We encode these equivalence classes using an
encoding similar to that for primitive propositions.  Let $\OpFalse,
\OpNot(\OpFalse),
\OpNot(\OpNot(\OpFalse)), \dots$ be an encoding of these 
equivalence classes, where we denote by $t_s$ the encoding of
$[s]_{\cK}$. Thus, $(s,s')\in\cK$ if and only if
$t_s=t_{s'}$. Construct the deductive algorithmic knowledge structure
$M'=(S',\pi',D)$, where $S'=\{(s,\{ t_s\})\mid s\in S\}$, and $\pi'$
is given as follows. For a term $t_p$ and state $s$,
$\pi'((s,\{t_s\}))(t_p)=\pi(s)(p)$. 
For all other terms $t$, we take
$\pi'((s,\{t_{s}\}))(t)=\false$.  It is  easy to check by induction on the
structure of $f$ that if $(M,s)\satS f$, then $(M',(s,\{ t_s
\}))\sat\hat{f}$. Here are the interesting cases of the induction. If
$f$ is a primitive proposition $p$, then by assumption, $(M,s)\satS p$, so $\pi(s)(p)=\true$;
thus, $\pi'((s,\{t_s\}))(t_p)=\true$, and $(M',(s,\{t_s\}))\sat
t_p$. If $f$ is $Kg$, then by assumption, $(M,s)\satS Kg$, so that for
all $s'\in\cK(s)$, $(M,s')\satS g$. By the induction hypothesis, we
have for all $s'\in\cK(s)$, $(M',(s',\{t_{s'}\}))\sat\hat{g}$, which
is equivalent to saying that for all $(s',\{t_{s'}\})\sim(s,\{t_s\})$
(since $t_{s'}=t_s$ exactly when $s'\in\cK(s)$), 
$(M',(s',\{t_{s'}\}))\sat\hat{g}$, and thus $(M',(s,\{t_s\}))\sat
K\hat{g}$, as required.

$(2)\rimp(3)$: This is immediate, since
$\cM_D(\Sigma)\subseteq\cM(\Sigma)$. 

$(3)\rimp(1)$: Assume $\hat{f}$ is satisfiable in a deductive
algorithmic knowledge structure $M=(S,\pi,D')$, that is,
$(M,s)\sat\hat{f}$ for some $s\in S$.  Construct the $\LK$-structure $M'=(S,\cK,\pi')$ by taking $\pi'(s)(p)=\true$ if
and only if $\pi(s)(t_p)=\true$,
and taking $\cK$ to be $\sim$. It is easy to check by induction on the structure of
$f$ that if $(M,s)\sat\hat{f}$, then $(M',s)\satS f$. Here are the interesting cases of the
induction. If $f$ is a primitive proposition $p$, then by assumption, $(M,s)\sat t_p$, so that 
$\pi(s)(t_p)=\true$. Thus means $\pi'(s)(p)=\true$, and thus
$(M',s)\satS p$. If $f$ is $Kg$, then by assumption, $(M,s)\sat
K\hat{g}$, that is, for all $s'\sim s$, $(M,s')\sat\hat{g}$. By the
induction hypothesis, this yields for all $s'\sim s$, $(M',s')\satS g$,
which is equivalent to the fact that for all $s'\in\cK(s)$,
$(M',s')\satS g$, that is, $(M',s)\satS Kg$, as required.
\end{proof}

Lemma~\ref{l:dedx1} says that we can relate the satisfiability of an
arbitrary formula of $\LK$ to that of a formula of $\LKDprop$. We can
similarly relate the satisfiability of an arbitrary formula of
$\LKDprop$ to that of a formula of $\LK$, in much the same
way. More precisely, given $\phi\in\LKDprop(\Sigma)$, let
$\tilde{\phi}$ be defined as follows.  The set $T^g_\Sigma$ is
countable, so let $\{p_t \mid t\in T^g_\Sigma\}$ be a countable set of
primitive propositions corresponding to the ground terms of
$T_\Sigma$.  Similarly, the set of formulas
$\{X\psi\mid\psi\in\LKDprop(\Sigma)\}$ is countable,
so let $\{q_\psi\mid \psi\in\LKDprop(\Sigma)\}$ be a
countable set of primitive propositions where $q_\psi$ corresponds to
the formula $X\psi$. Finally, let $\{r_t\mid t\in T^g_\Sigma\}$ be a
countable set of primitive propositions where $r_t$ corresponds to the
formula $\Ob(t)$.  Let $\tilde{\phi}$ be the translation of $\phi$
obtained by replacing every occurrence of a term $t$ in $T^g_\Sigma$
by $p_t$, every occurrence of a formula $X\psi$ by the corresponding
$q_\psi$, and every occurrence of a formula $\Ob(t)$ by the
corresponding $r_t$, in conjunction with
formulas $r_t\riff K r_t$ for all observations $\Ob(t)$ appearing
in $\phi$, formulas $q_\psi\riff Kq_\psi$ for all $X\psi$ appearing
in $\phi$, and formulas $r_t\rimp q_{\Ob(t)}$ for all observations
$\Ob(t)$ appearing in $\phi$. This translation is essentially
compositional: $\widetilde{\phi_1\land\phi_2}$ is logically equivalent
to $\tilde{\phi_1}\land\tilde{\phi_2}$, $\widetilde{\neg\phi}$ is
logically equivalent to $\neg\tilde{\phi}$, and $\widetilde{K\phi}$ is
logically equivalent to $K\tilde{\phi}$.  Note that
$\abs{\tilde{\phi}}$ is polynomial in $\abs{\phi}$.
\begin{lemma}\label{l:dedx2}
If $\phi\in\LKDprop(\Sigma)$, then $\phi$ is satisfiable in
$\cM(\Sigma)$ if and only if $\tilde{\phi}$ is satisfiable in an $\LK$-structure. 
\end{lemma}
\begin{proof}
Assume $\phi$ is satisfiable in $\cM(\Sigma)$, that is, there is a
deductive algorithmic knowledge structure $M=(S,\pi,D)$ such that
$(M,s)\sat\phi$ for some $s\in S$. Construct an $\LK$-structure
$M'=(S,\cK,\pi')$ by taking $\pi'(s)(p_t)=\pi(s)(t)$, 
$\pi'(s)(q_\psi)=\true$ if and only if $(M,s)\sat X\psi$,
$\phi'(s)(r_t)=\true$ if and only if $(M,s)\sat\Ob(t)$, and
$\cK$ is simply $\sim$. It is easy to check by induction on the structure of
$\phi$ that if $(M,s)\sat\phi$, then $(M',s)\satS\tilde{\phi}$. Here
are the interesting cases of the induction. If $\phi$ is a ground term
$t$, then by
assumption, $(M,s)\sat t$, and $\pi(s)(t)=\true$. This yields
$\pi'(s)(p_t)=\true$, and $(M',s)\satS p_t$. If $\phi$ is $X\psi$,
then by assumption, $(M,s)\sat X\psi$, and therefore
$\pi'(s)(q_\psi)=\true$, so that $(M',s)\satS q_\psi$. If $\phi$ is
$\Ob(p)$, then by assumption $(M,s)\sat\Ob(p)$, therefore
$\pi'(s)(r_p)=\true$, so that $(M',s)\satS r_p$. If $\phi$ is
$K\psi$, then by assumption, $(M,s)\sat K\psi$, that is, for all
$s'\sim s$, $(M,s')\sat \psi$. By the induction hypothesis, and the
definition of $\cK$, we have for all $s'\in\cK(s)$,
$(M',s')\satS\tilde{\psi}$, that is, $(M',s)\satS K\tilde{\psi}$, as
required.

Conversely, assume $\tilde{\phi}$ is satisfied in some $\LK$-structure.
By Theorem~\ref{p:ladner}, we know that there exists an $\LK$-structure
$M=(S,\cK,\pi)$ where  $\abs{S}\le\abs{\tilde{\phi}}$ and
$(M,s)\satS\tilde{\phi}$ for some $s\in S$. Let $\{ [s]_{\cK} \mid 
s\in S\}$ be the set of equivalence classes of $\cK$, of which there
are at most $\abs{\tilde{\phi}}$, which is polynomial in $\abs{\phi}$. Let
$t_1,t_2,\dots$ be an encoding of these
equivalence classes using terms $t_i\in T^g_\Sigma$ such that no
$\Ob(t_i)$ appears in $\phi$. We denote by $t_s$ the term encoding
the class 
$[s]_{\cK}$. Thus, $(s,s')\in\cK$ if and only if
$t_s=t_{s'}$. For every world $s\in S$, let $O(s)=\{p\mid 
\pi(s)(r_p)=\true,\text{$\Ob(p)$ appears in $\phi$}\}$. In
some sense, $O(s)$ represents the observations made at $s$. 
By the construction of $\tilde{\phi}$,  if $(s,s')\in\cK$, then
 $O(s)=O(s')$: assume $t\in O(s)$; then $(M,s)\sat \Ob(r_t)$,
and thus $(M,s)\sat K\Ob(r_t)$ (since $(M,s)\sat\tilde{\phi}$), so
that $(M,s')\sat\Ob(r_t)$, and $t\in O(s')$; the result follows by
symmetry of $\cK$. 
Construct the deductive algorithmic knowledge structure 
$M'=(S',\pi',D)$, where $S'=\{(s,\{ t_s \}\cup O(s))\mid s\in S\}$,
and $\pi'$ is obtained by taking 
$\pi'((s,O))(t)=\true$ if and only if $\pi(s)(p_t)=\true$. Finally,
take $D=\{ (\{t_s\},\psi^T)\mid
\pi(s)(q_\psi)=\true\}$. It is easy to check by
induction on the structure of $\phi$ that if $(M,s)\satS\tilde{\phi}$,
then $(M',(s,\{t_s\}\cup O(s)))\sat\phi$.  Here are the interesting
cases of the induction. If $\phi$ is a term $t$, then by assumption,
$(M,s)\satS p_t$, so that $\pi(s)(p_t)=\true$. Thus, we have
$\pi'((s,\{t_s\}\cup O(s)))(t)=\true$, and
$(M',(s,\{t_s\}\cup O(s)))\sat t$. If $\phi$ is $X\psi$, then we
have by assumption $(M,s)\satS q_\psi$, and so $\pi(s)(q_\psi)=\true$,
meaning that $(\{t_s\},\psi^T)$ is a deduction rule in $D$, and thus
$(M',(s,\{t_s\}\cup O(s)))\sat X\psi$. If $\phi$ is $\Ob(t)$, then
by assumption $(M,s)\sat r_t$, so that $\pi(s)(r_t)=\true$. Thus,
$t\in O(s)$, and therefore
$(M',(s,\{t_s\}\cup O(s)))\sat\Ob(t)$. If $\phi$ is $K\psi$,
consider an arbitrary $s'$ such that
$(s',\{t_{s'}\}\cup O(s'))\sim(s,\{t_s\}\cup O(s))$. This
certainly implies, by the assumptions on the encoding, that
$t_s=t_{s'}$, and thus $s'\in\cK(s)$. By the fact that $(M,s)\sat
K\tilde{\psi}$, we have $(M,s')\sat\tilde{\psi}$, and by the induction
hypothesis, $(M',\{t_{s'}\}\cup O(s'))\sat\psi$. Since $s'$ was
arbitrary, we get $(M',\{t_s\}\cup O(s))\sat K\psi$, as required.
\end{proof}

\begin{oldtheorem}{t:decision1}
\begin{theorem}
The problem of deciding whether a formula $\phi$ of $\LKDprop(\Sigma)$
is satisfiable  in $\cM(\Sigma)$ is NP-complete.
\end{theorem}
\end{oldtheorem}
\begin{proof}
For the lower bound, we show how to reduce from the decision problem
of $\LK$. Let $f$ be a formula of $\LK$. By Lemma~\ref{l:dedx1}, $f$ is 
satisfiable if and only if $\hat{f}$ is satisfiable in
$\cM(\Sigma)$. Thus, the complexity of the decision problem for $\LK$ is
a lower bound for our decidability problem, that is, NP. 
For the upper bound, we need to exhibit a nondeterministic polynomial
time algorithm that decides if $\phi\in\LKDprop(\Sigma)$ is
satisfiable. We will use the decision problem for $\LK$ itself as an
algorithm. By Lemma~\ref{l:dedx2}, $\phi$ is 
satisfiable if and only if $\tilde{\phi}$ is satisfiable, so we can
simply invoke the NP algorithm for $\LK$ satisfiability on $\tilde{\phi}$. 
\end{proof}

\begin{oldtheorem}{t:lowerbound}
\begin{theorem}
For any given $\SigmaKD$-deductive system $D$, the problem of
deciding whether a formula $\phi$ of $\LKDprop(\Sigma)$ is satisfiable
in $\cM_D(\Sigma)$ is NP-hard.
\end{theorem}
\end{oldtheorem}
\begin{proof}
The lower bound follows from Lemma~\ref{l:dedx1}. Let $f$ be an $\LK$ 
formula; $f$ is satisfiable if and only if $\hat{f}$ is
satisfiable over $\cM_D(\Sigma)$ structures. Since the decision
problem for $\LK$ is NP-complete, the lower bound follows. 
\end{proof}

The following small model result for $\LKDprop(\Sigma)$ over
$\cM^n_D(\Sigma)$ is needed in the proof of
Theorem~\ref{t:decision}. Define the size $\abs{M}$ of a model
$M$ to
be the sum of the sizes of the states, where the size of a state
$(e,\{p_1,\dots,p_k\})$ is $1+\abs{p_1}+\dots+\abs{p_k}$.

\begin{lemma}\label{l:small-model}
Let $P(x)$ be a polynomial. If $\phi$ is satisfiable in
$M\in\cM^{P(\abs{\phi})}_D(\Sigma)$, then $\phi$ is 
satisfiable in a structure $M'\in\cM^{P(\abs{\phi})}_D(\Sigma)$ with
$\abs{M'}$ polynomial in $\abs{\phi}$. 
\end{lemma}
\begin{proof}
Assume $\phi$ is satisfiable in some structure $M$. Let
$M_1=(S_1,\cK_1,\pi_1)$ be the $\LK$-structure obtained by the
construction in Lemma~\ref{l:dedx2}, with
$(M_1,s_1)\satS\tilde{\phi}$, for some $s_1=(e,O)$ in $S_1$.
By Lemma~\ref{p:ladner},
we know that $\tilde{\phi}$ is satisfied in an
$\LK$-structure $M_2=(S_2,\cK_2,\pi_2)$ where $\abs{S_2}\le\abs{\tilde{\phi}}$,
$\cK_2$ is a universal relation on $S_2$ (that is, $\cK_2=S_2\times
S_2$), and $(M_2,s_2)\satS\tilde{\phi}$ for some $s_2\in S_2$. 
We reconstruct a satisfying deductive algorithmic knowledge structure
from $M_2$. Specifically, define $M'=(S',\pi',D)$ by taking $S'=\{
(s,O) \mid s\in S_2\}$ (where $O$ is the set of observations
at state $s_1$), and $\pi'((s,O))(t)=\pi_2(s)(p_t)$.
Because the construction of Lemma~\ref{l:dedx2} does not change the
number of observations at a state and $M\in\cM^{P(\abs{\phi})}$, we
have $\abs{s_1}\le P(\abs{\phi})$, and thus $\sum_{p\in O}\abs{p}\le
P(\abs{\phi})$. Thus, $|M'|$ is polynomial in
$\abs{S_2}P(\abs{\phi})\le \abs{\tilde{\phi}}P(\abs{\phi})$, that is, polynomial in 
$\abs{\phi}$, as required.
A straightforward induction on the structure of 
$\phi$ shows that if $(M,s)\sat\phi$ (or equivalently, by 
Lemma~\ref{l:dedx2}, $(M_1,s_1)\satS\tilde{\phi}$ for some $s_1$), then 
$(M',(s_2,O))\sat\phi$, for some $s_2$. Here are the interesting cases of the
induction. 
If $\phi$ is a ground term $t$, then $(M_1,s_1)\sat p_t$, and
$\pi_1(s_1)(p_t)=\true$; this means that $\pi_2(s_2)(p_t)=\true$ (by
construction of $M_2$), so that $\pi'((s_2,O))(t)=\true$, and
$(M',(s_2,O))\sat t$. If $\phi$ is $X\psi$, then by the fact that
$(M,s)\sat X\psi$, and that $s=(e,O)$, we have $\{\OpOb(p)\mid p\in O\}\vdash_D
\psi^T$, and thus, $(M',(s_2,O))\sat X\psi$, since the same
observations are used at $s_2$. Similarly, if $\phi$ is $\Ob(t)$, then
$(M,s)\sat\Ob(t)$, and if $s=(e,O)$, then $t\in O$, and thus
$(M',(s_2,O))\sat\Ob(t)$, since the same observations are used at
$s_2$. Finally, if $\phi$ is $K\psi$, then consider
an arbitrary $s'$ such that $(s',O)\sim(s_2,O)$; since all
states have the same observations, $s'$ can be arbitrary in $S_2$. Since
$\cK_2$ was the universal relation on $S_2$, we have
$s'\in\cK_2(s_2)$. By assumption, we know $(M_1,s_1)\satS
K\tilde{\psi}$, and thus $(M_2,w_2)\satS K\tilde{\psi}$, so that
$(M_2,s')\satS \tilde{\psi}$. By the induction hypothesis,
$(M',(s',O))\sat\psi$, and since $s'$ was arbitrary,
$(M',(s_2,O))\sat K\psi$, as required.
\end{proof}

\begin{oldtheorem}{t:decision}
\begin{theorem}
For any given $\SigmaKD$-deductive system $D$ that is decidable in
nondeterministic polynomial time and for any polynomial $P(x)$, the
problem of deciding 
whether a formula $\phi$ of $\LKDprop(\Sigma)$ is satisfiable in
$\cM^{P(\abs{\phi})}_D(\Sigma)$ is NP-complete. 
\end{theorem}
\end{oldtheorem}
\begin{proof}
The lower bound is given by Theorem~\ref{t:lowerbound}.  For the upper
bound, we can do something similar to what we did in
Theorem~\ref{t:decision1}, except we need to keep track of the size of
the objects we manipulate. Let $\phi$ be a formula of
$\LKDprop(\Sigma)$. We exhibit an algorithm that nondeterministically
decides if $\phi$ is satisfiable. From Lemma~\ref{l:small-model}, it
suffices to nondeterministically guess a satisfying structure $M$ with
a set of worlds polynomial in $\abs{\phi}$, which is guaranteed to
exist if and only if $\phi$ is satisfiable. Moreover, for every
subformula $X\psi$ of $\phi$ (of which there are polynomially many)
and every state $s$ of $M$ (of which there are polynomially 
many), we nondeterministically guess
whether the observations at $s$ (of which there are polynomially many)
can derive $\psi^T$. We can verify that $\phi$
is satisfied in $M$ in time polynomial in $\abs{\phi}$, by adapting
the polynomial time algorithm of \cite[Proposition
3.1]{r:halpern92}. Roughly speaking, the algorithm consists of
enumerating all the subformulas of $\phi$, and for each subformula
$\psi$ (in order of length), marking every state of $M$ with either
$\psi$ or $\neg\psi$ depending on whether $\psi$ or $\neg\psi$ holds
at the state: primitive propositions are handled by invoking the
interpretation, formulas of the form $X\psi'$ are handled by verifying
if the guess of whether $\psi'^T$ is derivable from the observations
at the state is correct, formulas of the form $\Ob(p)$ are handled by looking up
$p$ in the observations at the state (the number of which is
polynomial in $\abs{\phi}$), conjunctions and negations are handled in
the obvious way, and formulas $K\psi'$ are handled by looking up
whether every reachable state from the current state is marked with
$\psi'$.
\end{proof}

\section{Proofs for Section~\ref{s:multiagent}}\label{a:section6}

\begin{oldtheorem}{t:another-axiomatization}
\begin{theorem}
The axiomatization $\AX_n$ is sound and complete for
$\LKDnprop(\Sigma)$ with respect to $\cM_n(\Sigma)$.  
\end{theorem}
\end{oldtheorem}
\begin{proof}
This is a  generalization of the proof of
Theorem~\ref{t:soundcomplete1}. Soundness is easy to check. For
completeness, we again show that if $\phi$ is consistent, 
then $\phi$ is satisfiable. We give the definitions here, leaving the
details of the proof to the reader. Given a set $V$ of formulas, let
$V/K_i=\{\phi\mid K_i\phi\in V\}$. Let $\C$ be the set of all maximal
consistent sets of formulas of $\LKDnprop(\Sigma)$. 
For $V\in\C$, let $V/\Ob_i=\{p\mid \Ob_i(p)\in V\}$. We define
$\approx_i$ over $\C$, for every $i$, by taking $V\approx_i U$ if and
only if $V/K_i\subseteq U$. We can check that $\approx_i$ is 
an equivalence relation for every $i$, assuming the axioms
\axiom{K1}--\axiom{K5}, just like in the proof of
Lemma~\ref{l:approx-props}. 
We can also check that if $V\approx_i U$, then $V/\Ob_i=U/\Ob_i$, and that
for all $\psi$, if $V\approx_i U$, then $X_i\psi\in V$ if and only if
$X_i\psi\in U$. 

As mentioned, this is a generalization of the proof of
Theorem~\ref{t:soundcomplete1}, in that the satisfying model is built
from maximally consistent sets of formulas. However, it is not simply
a direct generalization. For Theorem~\ref{t:soundcomplete1}, it was
sufficient to consider a single equivalence class of the relation
$\approx$ as the set of states: all the states could be assumed to
have the same observations, thus $\sim$ could be taken to be a
universal relation in the canonical model. That this can be done is
strongly related to Theorem~\ref{p:ladner}, which says that if a
formula of $\LK$ is satisfiable at all (in an $\LK$-structure), it is
satisfiable in a structure with a universal relation. That result does
not hold, however, when we consider multiple agents. This makes the
argument slightly more complex.

Let $\phi$ be a consistent formula of $\LKDnprop(\Sigma)$,
and let $\mathit{Sub}(\phi)$ be the set of subformulas of $\phi$
(including $\phi$ itself). Let $O^\phi = \{p\mid
\text{$\Ob_i(p)\in\mathit{Sub}(\phi)$, for some $i$}\}$, 
$X^\phi=\{\psi\mid \text{$X_i\psi\in\mathit{Sub}(\phi)$, for some
$i$}\}$, and
$K^\phi=\{\psi\mid\text{$K_i\psi\in\mathit{Sub}(\phi)$}\}$. Clearly,
$O^\phi$, $X^\phi$, and $K^\phi$ are finite sets. First, we 
claim that any consistent set of formulas $F$ (with
$F/\Ob_i\subseteq O^\phi$ for all $i$) can be extended to a
maximally consistent set $F'$ (with $F'/\Ob_i\subseteq O^\phi$ for
all $i$): construct the set $F''$ incrementally starting with $F$,
adding $\Ob(p)$ for every observation $p\in O^\phi$ if $\Ob(p)$ is
consistent with the current set, and adding $\neg\Ob(p)$ for every
observation $p$ either not appearing in $O^\phi$ or inconsistent with
the current set; it is easy to establish that $F''$ is consistent, so
$F''$ is extensible to a maximally consistent set $F'$ with
$F'/\Ob_i\subseteq O^\phi$ for all $i$. Let $\C(\phi)$ be the set of
all maximally consistent sets of formulas $F$ with $F/\Ob_i\subseteq
O^\phi$ for all $i$. We shall use $\C(\phi)$ as our states. The fact
that we consider $\C(\phi)$ means, roughly, that we consider only
observations in $O^\phi$ as relevant. 

The set $X^\phi$ is finite, so let $S_1,\dots,S_{2^{|X^\phi|}}$ be an
enumeration of the subsets of $X^\phi$. Let
$p_1,\dots,p_{2^{|X^\phi|}}$ be a set of primitive propositions
not in $O^\phi$, where we associate $p_i$ with $S_i$. Define
the function $\mi{tag}_X(V)$ mapping every set $V\in\C(\phi)$ to the
primitive proposition corresponding to the set $(\cup_i(V/X_i)) \cap
X^\phi$ where $V/X_i=\{\psi\mid X_i\psi\in V\}$. Thus, $\mi{tag}_X(V)$
gives the primitive proposition corresponding to the formulas in
$X^\phi$ that appear under an $X_i$ in $V$. 

In a similar way, the set $K^\phi$ is finite, so let
$T_1,\dots,T_{2^{|K^\phi|}}$ be an 
enumeration of the subsets of $K^\phi$. Let
$q_1,\dots,q_{2^{|X^\phi|}}$ be a set of primitive propositions
not in $O^\phi$ or $\{p_1,\dots,p_{2^{|X^\phi|}}\}$, where we
associate $q_i$ with $T_i$. Define 
the function $\mi{tag}_K(V)$ mapping every set $V\in\C(\phi)$ to the
primitive proposition corresponding to the set $(\cup_i(V/K_i)) \cap
K^\phi$. Thus, $\mi{tag}_K(V)$
gives the primitive proposition corresponding to the formulas in
$K^\phi$ that appear under an $K_i$ in $V$. 

Since $\phi$ is consistent, there is a set $V^\phi\in 
\C(\phi)$ with $\phi\in V^\phi$.
Define the deductive algorithmic knowledge structure
$M^\phi=(S^\phi,\pi^\phi,D_1^\phi,\dots,D_n^\phi)$ by taking
\begin{align*}
 & S^\phi  = \{ 
(s_V,\begin{prog}\{\mathit{tag}_X(V),\mathit{tag}_K(V)\}\cup
V/\Ob_1,\dots,\\
  \{\mathit{tag}_X(V),\mathit{tag}_K(V)\}\cup V/\Ob_n)\mid
 V\in\C(\phi)\}\end{prog}\\
 & \pi^\phi((s_V,O_1,\dots,O_n))(p)  = \begin{cases}
                  \true & \text{if $p\in V$}\\
                  \false & \text{if $p\not\in V$}
                  \end{cases}\\
 & D_i^\phi  = \{ \begin{prog}
  (\{\OpOb_i(\mi{tag}_X(V))\}\cup V/\Ob_i,\psi^T) \mid \\
  \qquad 
  V\in \C(\phi),
  X_i\psi\in\mathit{Sub}(\phi), X_i\psi\in V\}. \end{prog}
\end{align*}
We can check that $M^\phi$ is a deductive algorithmic knowledge
structure with $n$ agents. Moreover, it is easy to check that if
$s_V\sim_i s_U$ in $M^\phi$, then for all $\psi\in X^\phi$,
$X_i\psi\in V$ if and only if $X_i\psi\in U$. (Indeed, if $s_V\sim_i
s_U$, then $\mi{tag}_X(V)=\mi{tag}_X(U)$, and the result follows from
the 
choice of tags.) Clearly, we also have that if $V\approx_i U$, then
$s_V\sim_i s_U$: we already know that $V\approx_i U$ implies
$V/\Ob_i=U/\Ob_i$, and $V\approx_i U$ means that the same
$X_i\psi$ formulas are in $V$ and $U$, and therefore, we also have
$\mi{tag}_X(V)=\mi{tag}_X(U)$, and similarly the same $K_i\psi$
formulas are in $V$ and $U$ and $\mi{tag}_K(V)=\mi{tag}_K(U)$, and
thus $s_V\sim_i s_U$.  Moreover, we have that if $s_U\sim_i s_V$, then
the same $K_i\phi$ are formulas are in $U$ and $V$, since
$\mi{tag}_K(U)=\mi{tag}_K(V)$. 

We can prove, adapting the proof of Theorem~\ref{t:soundcomplete1},
that for all $s_V\in S^\phi$ and all subformulas
$\psi\in\mathit{Sub}(\phi)$, $(M^\phi,s_V)\sat\psi$ if and only if
$\psi\in V$.  Here are the interesting cases of the induction. 

For $\Ob_i(p)$, first assume that $(M^\phi,s_V)\sat\Ob_i(p)$. This means
that $p\in V/\Ob_i$ (since $\mi{tag}_X(V),\mi{tag}_K(V)\not\in
O^\phi$, we cannot 
have $p=\mi{tag}_X(V)$ or $p=\mi{tag}_K(V)$), and thus $\Ob_i(p)\in
V$. Conversely, if $\Ob_i(p)\in V$, then $p\in V/Ob_i$, so
that $(M^\phi,s_V)\sat\Ob_i(p)$.

Now, consider a deductive algorithmic knowledge formula
$X\psi$. First, assume that we have $(\uMp,s_V)\sat X_i\psi$.  By
definition, $\{\OpOb_i(\mi{tag}_X(V)),\OpOb_i(\mi{tag}_K(V))\}\cup\{\OpOb_i(p)\mid
p\in V/\Ob_i\}\vdash_{D_i^\phi}\psi^T$. If $\psi$ is $\Ob_i(p)$ for
some $p\in V/\Ob_i$ (we cannot have $p=\mi{tag}_X(V)$ or $p=\mi{tag}_K(V)$, by choice of
tags), then $(M^\phi,s_V)\sat\Ob_i(p)$, and thus $\Ob_i(p)\in V$ by
the induction hypothesis; by \axiom{X2}, $X_i\Ob_i(p)\in
V$. Otherwise, consider the derivation of $\psi^T$. By examination of
$D_i^\phi$, the last rule in this derivation must have
$\OpOb_i(\mi{tag}_X(V))$ in the premise (since there is a single tag in the
premise of every rule, and the tag at 
state $V$ is $\mi{tag}_X(V)$). By definition of $D_i^\phi$, this means
that $X_i\psi\in V$, as required.  Conversely,
assume that $X_i\psi\in V$. By definition of $D_i^\phi$,
$(\{\OpOb_i(\mi{tag}_X(V))\}\cup\{\OpOb_i(p)\mid
p\in V/\Ob_i\},\psi^T)\in D_i^\phi$, and thus
$\{\OpOb_i(\mi{tag}_X(V))\}\cup\{\OpOb_i(p)\mid
p\in V/\Ob_i\}\vdash_{D_i^\phi}\psi^T$, meaning that $(\uMp,s_V)\sat 
X_i\psi$.

For a knowledge formula $K_i\psi$, the argument is similar to that in
the proof of Theorem~\ref{t:soundcomplete1}.

Completeness follows 
from the fact that $\phi\in V^\phi$
and $\phi\in\mathit{Sub}(\phi)$, so that $(\uMp,s_V)\sat\phi$, and
thus $\phi$ is satisfiable.
\end{proof}

\begin{oldtheorem}{t:another-axiomatization2}
\begin{theorem}
The axiomatization
$\AX_n$ augmented with axioms $\Axprop^{D_1}_n,\dots,\Axprop^{D_n}_n$
is sound and complete for $\LKDnprop(\Sigma)$ with respect to
$\cM_{D_1,\dots,D_n\subseteq}(\Sigma)$.   
\end{theorem}
\end{oldtheorem}
\begin{proof}
Soundness is again straightforward. 
For completeness, we prove
the equivalent statement that if $\phi$ is consistent then $\phi$ is satisfiable in
some structure in $\cM_{D_1,\dots,D_n}(\Sigma)$. The procedure 
is exactly the one that is used to prove
Theorem~\ref{t:another-axiomatization}, except that we construct
the deductive systems $D_1^\phi,\dots,D_n^\phi$ differently.

We simply indicate where the proof differs from that of
Theorem~\ref{t:another-axiomatization}, and let the reader fill in the
details. We construct, for a given $\phi$, a deductive algorithmic
knowledge structure with $n$ agents
$M^\phi=(S^\phi,\pi^\phi,D^\phi_1,\dots,D^\phi_n)$, where $S^\phi$ and
$\pi^\phi$ are constructed as in 
Theorem~\ref{t:another-axiomatization}, and $D^\phi_1,\dots,D^\phi_n$
are obtained by taking $D^\phi_i$ to be
\[ D_i \cup \{ \begin{prog}
  (\{\OpOb(\mi{tag}_X(V))\}\cup V/\Ob_i,\psi^T) \mid \\
  \qquad V\in\C(\phi), X_i\psi\in\mathit{Sub}(\phi), 
  X_i\psi\in V\}. \end{prog}\]
$M^\phi$ is a deductive algorithmic knowledge
structure in $\cM_{D_1,\dots,D_n\subseteq}(\Sigma)$. As in the
proof of Theorem~\ref{t:soundcomplete3}, we can show that 
if $\{\OpOb_i(\mi{tag}_X(V))\}\cup\{\OpOb_i(p)\mid
p\in V/\Ob_i\}\vdash_{D_i}\psi^T$, for some $\psi\in X^\phi$, then
there is a rule  $\{\OpOb_i(\mi{tag}_X(V))\}\cup\{\OpOb_i(p)\mid
p\in V/\Ob_i\}\triangleright\psi^T$ in $D_i^\phi$. The argument is
quite similar, by showing that if $\{\OpOb_i(\mi{tag}_X(V))\}\cup\{\OpOb_i(p)\mid
p\in V/\Ob_i\}\vdash_{D_i}\psi^T$, then $X_i\psi$ must be in $V$.
Once this is established, the rest of the proof follows that of
Theorem~\ref{t:another-axiomatization}.  
\end{proof}

\begin{oldtheorem}{t:complexity-multiagent}
\begin{theorem}
If $n\ge 2$, the problem of deciding whether a formula $\phi$ of
$\LKDprop_n(\Sigma)$ is satisfiable in $\cM_n(\Sigma)$ is
PSPACE-complete. 
\end{theorem}
\end{oldtheorem}
\begin{proof}
The proof is entirely analogous to that of Theorem~\ref{t:decision1},
except that we use the modal logic $\LKn$ 
rather than $\LK$. We can define translations between 
$\LKDnprop(\Sigma)$ and $\LKn$, and we can prove analogues of 
Lemmas~\ref{l:dedx1} and \ref{l:dedx2}.  We simply give the translations
here, leaving the reader to fill in the details. 

Let $f$ be a formula of $\LKn$. Let $p_1,\ldots,p_k$ be the primitive
propositions appearing in $f$. We first come up with an encoding of
these primitive propositions into the language of $\Sigma$. For
example, we can take $p_1$ to be $\OpTrue$, $p_2$ to be
$\OpNot(\OpTrue)$, $p_3$ to be $\OpNot(\OpNot(\OpTrue))$, and so
forth. Let $t_p$ be the term encoding the primitive proposition
$p$.  Let $\hat{f}$ be the formula
obtained by replacing every instance of a primitive proposition $p$ in
$f$ by $t_p$. Note that $\abs{\hat{f}}$ is polynomial in $\abs{f}$, and that
$\hat{f}$ contains no instance of the $X$ operator.  We can show that
$f$ is satisfiable in $\LKn$-structures for $n$ agents if and only
if $\hat{f}$ is satisfiable in $\cM_n(\Sigma)$, with a proof similar to that of
Lemma~\ref{l:dedx1}. This gives us an immediate lower bound, as
follows. Let $f$ be an $\LKn$ formula. We know $f$ is satisfiable if
and only if $\hat{f}$ is satisfiable over $\cM_n(\Sigma)$
structures. Since the decision problem for $\LKn$ ($n\ge 2$) is
PSPACE-complete, the lower bound of PSPACE follows.

Let $\phi$ be a formula of $\LKDnprop(\Sigma)$.  The set $T^g_\Sigma$
is countable, so let $\{p_t \mid t\in T^g_\Sigma\}$ be a countable set
of primitive propositions corresponding to the ground terms of
$T_\Sigma$.  Similarly, for every $i$, the set of formulas
$\{X_i\psi\mid\psi\in\LKDnprop(\Sigma)\}$ is
countable, so let $\{q^i_\psi\mid
\psi\in\LKDnprop(\Sigma)\}$ be a countable set of
primitive propositions where $q^i_\psi$ corresponds to the formula
$X_i\psi$. Finally, let $\{r^i_t\mid t\in T^g_\Sigma\}$ be a countable
set of primitive propositions where $r^i_t$ corresponds to the formula
$\Ob_i(t)$. Let $\tilde{\phi}$ be the translation of $\phi$ obtained
by replacing every occurrence of a term $t$ in $T^g_\Sigma$ by $p_t$,
every occurrence of a formula $X_i\psi$ by the corresponding
$q^i_\psi$, and every occurrence of a formula $\Ob_i(t)$ by the
corresponding $r^i_r$, in conjunction
with formulas $r^i_t\riff K_i r^i_r$ for all observations
$\Ob_i(t)$ appearing in $\phi$, formulas $q^i_\psi\riff K_i
q^i_\psi$ for all formulas $X_i\psi$ appearing in $\phi$, and formulas
$r^i_t\rimp q^i_{\Ob_i(t)}$ for all observations $\Ob_i(t)$ appearing
in $\phi$. Note that $\abs{\tilde{\phi}}$ is polynomial in
$\abs{\phi}$.  We can show that $\phi$ is satisfiable in
$\cM_n(\Sigma)$ if and only if $\tilde{\phi}$ is satisfiable in an
$\LKn$-structure, using a proof similar to that of
Lemma~\ref{l:dedx2}. This gives us an immediate upper bound for our
decision problem: $\phi$ is satisfiable if and only if $\tilde{\phi}$
is satisfiable, so we can simply invoke the PSPACE algorithm for
$\LKn$ satisfiability on $\tilde{\phi}$.
\end{proof}

\bibliographystyle{chicagor}
\bibliography{riccardo2,joe,z,refs}

\end{document}